\documentclass[twoside,english,3p]{elsarticle}
\usepackage[T1]{fontenc}
\usepackage{geometry}
\usepackage{amssymb}
\usepackage{amsmath}

\usepackage{amsthm}
\usepackage{stmaryrd}
\usepackage{graphicx}
\usepackage{booktabs}
\usepackage{multirow}
\usepackage{makecell}
\usepackage{xurl}
\usepackage{esint}
\usepackage{algorithm}
\usepackage{algpseudocode}
\usepackage{rotating}
\usepackage{adjustbox}
\usepackage{nomencl}
\usepackage{multicol}
\makenomenclature
\makeatletter
\theoremstyle{plain}

\theoremstyle{boldremark} 

\ifx\proof\undefined

\providecommand{\proofname}{Proof}
\fi

\renewenvironment{proof}[1][\proofname]{%
	\par\pushQED{\qed}\normalfont%
	\topsep6\p@\@plus6\p@\relax
	\trivlist\item[\hskip\labelsep\bfseries#1\@addpunct{.}]%
	\ignorespaces
}{%
	\popQED\endtrivlist\@endpefalse
}

\journal{Elsevier}

\usepackage{hyperref}
\hypersetup{colorlinks = true, allcolors = blue}

\usepackage[nameinlink]{cleveref}

\Crefname{figure}{Fig.}{Figs.}
\Crefformat{equation}{Eq.~#2(#1)#3}
\Crefformat{section}{Section~#2#1#3}
\AtBeginDocument{%
	\let\citet\cite
}

\usepackage[labelfont=bf]{caption}
\@ifundefined{showcaptionsetup}{}{%
	\PassOptionsToPackage{caption=false}{subfig}}
\usepackage{subfig}
\makeatother

\usepackage{babel}
\providecommand{\remarkname}{Remark}
\providecommand{\theoremname}{Theorem}

\begin{document}
	
	\begin{frontmatter}{}
		
		\title{Replay-Based Continual Learning for Physics-Informed Neural Operators}

\author[rvt,rvt3]{Yizheng Wang}

\ead{wang-yz19@tsinghua.org.cn}

\author[rvt6]{Mohammad Sadegh Eshaghi}

\author[rvt6]{Xiaoying Zhuang}

\author[rvt3]{Timon Rabczuk\corref{cor1}}

\author[rvt]{Yinghua Liu\corref{cor1}}

\ead{yhliu@mail.tsinghua.edu.cn}
\cortext[cor1]{Corresponding author}
\address[rvt]{Department of Engineering Mechanics, Tsinghua University, Beijing 100084, China}

\address[rvt3]{Institute of Structural Mechanics, Bauhaus-Universit\"{a}t Weimar, Marienstr. 15, D-99423 Weimar, Germany}	

\address[rvt6]{ Institute of Photonics, Department of Mathematics and Physics, Leibniz University Hannover, Germany}

\begin{abstract}

Neural operators generally demonstrate strong predictive performance on in-distribution (ID) problems. However, a critical limitation of existing methods is their significant performance degradation when encountering out-of-distribution (OOD) data. To address this issue, this work introduces continual learning into physics-informed neural operators, with particular emphasis on neural operators built upon the Transolver architecture, and proposes a simple yet effective replay-based continual learning strategy.
The proposed method is fully physics-informed and does not require labeled data, relying solely on input fields together with physical constraints for training. When new OOD data become available, a small number of past data are incorporated through a distillation-based constraint to preserve previously acquired knowledge and alleviate catastrophic forgetting. Meanwhile, a transfer learning LoRA is employed to enable rapid adaptation to the new data.
The proposed framework is systematically validated on three representative physical problems, including the Darcy flow problem in fluid mechanics, a two-dimensional hyperelastic brain tumor problem in biomechanics, and a three-dimensional linear elastic Triply Periodic Minimal Surfaces problem in solid mechanics. The results demonstrate that the proposed method effectively mitigates catastrophic forgetting on previously learned data while maintaining fast adaptability to new data. Compared with conventional joint training strategies, the proposed method significantly improves training efficiency while reducing additional memory usage and computational cost.

\end{abstract}

\printnomenclature

		
\end{frontmatter}{}

\section{Introduction}

Partial differential equations (PDEs) are fundamental for understanding and predicting physical processes. Traditional numerical methods typically require solving each new problem instance independently, meaning that whenever material distributions, geometries, or boundary conditions change, the governing equations must be solved again from scratch \citet{wang2021learning}. For complex large-scale problems, such repeated computations often incur prohibitively high computational cost.

In recent years, neural operators have emerged as a promising paradigm for the rapid solution of PDEs \citet{wang2023scientific}. By directly learning mappings from input function spaces to output solution spaces, neural operators can efficiently predict target solutions even when material parameters, geometries, or boundary conditions vary. Compared with conventional numerical solvers, neural operators often achieve speedups of several orders of magnitude, thereby significantly reducing online computational cost \citet{kovachki2023neural}. Representative neural operator architectures currently include DeepONet \citet{DeepOnet}, FNO \citet{li2020fourier}, and transformer-based neural operators \citet{hao2023gnot,wu2024transolver}. Although existing neural operators generally perform well on in-distribution (ID) problems, a critical limitation remains their poor generalization under out-of-distribution (OOD) scenarios. OOD problems refer to situations where the training and testing data follow different distributions, leading to substantial deterioration in predictive accuracy.

To partially address the OOD issue, continual learning provides a promising solution \citet{wang2024comprehensive}. It is well recognized that the ability to learn tasks in a sequential fashion is crucial to the development of artificial intelligence \citet{kirkpatrick2017overcoming}. Continual learning refers to a learning paradigm in which a model continuously acquires knowledge from a sequence of tasks while attempting to preserve previously learned information and mitigate catastrophic forgetting \citet{wang2024comprehensive}. This capability is particularly important in practical applications because data from different distributions often arrive sequentially over time. Consequently, models must be periodically updated to absorb newly available data, thereby progressively improving generalization under evolving distributions. Since neural operators also suffer from OOD degradation, exploring continual learning strategies for neural operators is of significant importance \citet{tripura2023foundational,elhadidy2026sle}. With the continuous accumulation of data, continual learning is expected to play a central role in future scientific foundation models \citet{menon2026scientific,choi2025defining}, especially in the emerging context of pretrained large-scale physics models \citet{hao2024dpot,zhou2024unisolver,mccabe2025walrus}. However, research on continual learning for neural operators remains relatively limited. Below we review some representative works. Tripura et al. \citet{tripura2023foundational} proposed a continual transfer framework based on modular neural operators, where shared experts are frozen and only gate parameters are updated to adapt to new PDE tasks. Nevertheless, when new tasks differ substantially from previously learned tasks, relying solely on gate recombination often leads to limited adaptability. Moreover, Elhadidy et al. \citet{elhadidy2026sle} introduced continual learning into FNO by expanding model parameters; however, their approach remains purely data-driven and does not incorporate physical constraints from PDEs.

Transformer-based neural operators have recently demonstrated remarkable potential in complex problems, such as multiphysics weather forecasting \citet{bi2023accurate} and automotive problems involving complex geometries \citet{wu2024transolver}. However, these transformer-based neural operators are typically trained in a fully data-driven manner. More recently, physics-driven neural operator training has shown significant promise, particularly in PFEM \citet{wang2026pfem} and VINO \citet{eshaghi2025variational}. Since PDE-driven training avoids expensive data generation from conventional numerical solvers, it offers substantial advantages in training efficiency. With the rapid development of scientific machine learning, scientific foundation models are expected to emerge in the near future \citet{choi2025defining,menon2026scientific}. In particular, transformer-based neural operators trained directly from governing physical equations are likely to become a key enabling technology for future physics foundation models. Because future foundation models will inevitably encounter continuously evolving datasets, exploring continual learning for transformer-based neural operators trained directly from PDEs becomes critically important. To the best of our knowledge, however, no existing work has systematically investigated this direction.

Therefore, this work presents a systematic study of continual learning for transformer-based physics-informed neural operators trained directly from governing equations. In particular, we propose a simple yet efficient replay-based continual learning strategy that focuses on a small number of poorly performing samples, thereby improving training efficiency while reducing memory usage. The main characteristics of the proposed replay-based continual learning framework are summarized as follows:

\begin{itemize}
	\item \textbf{Pure physics-driven training}: We train a Transolver-based physics-informed neural operator directly from governing equations, completely without relying on labeled data.
	
	\item \textbf{Focusing on poorly performing samples}: A PDE-based scoring strategy is introduced to rank all samples, and poorly performing samples are selectively replayed to improve training efficiency while reducing memory cost.
	
	\item \textbf{Mitigation of catastrophic forgetting}: We exploit the old model via distillation to preserve performance on past data, thereby mitigating catastrophic forgetting.
	
	\item \textbf{Adaptation to new data}: The model can adapt well to newly introduced distributions, absorbing new data and partially alleviating the OOD problem.
	
	\item \textbf{Supervised fine-tuning}: Inspired by supervised fine-tuning in foundation models, we incorporate this technique into neural operator training to further improve predictive accuracy.
\end{itemize}

The proposed replay-based continual learning framework is systematically validated on three representative physical problems, including the Darcy flow problem in fluid mechanics, a two-dimensional hyperelastic brain tumor problem in biomechanics, and a three-dimensional linear elastic Triply Periodic Minimal Surfaces (TPMS) problem in solid mechanics \citet{al2019multifunctional}. The results demonstrate that the replay-based continual learning approach effectively mitigates catastrophic forgetting on past data while maintaining strong adaptability to new data. Moreover, substantial improvements in training efficiency are achieved.

The outline of the paper is as follows. \Cref{sec:Preparatory-knowledge} introduces neural operators and continual learning. \Cref{sec:Method} classifies the OOD scenarios encountered by neural operators and presents our proposed replay-based continual learning together with the supervised fine-tuning (SFT) strategy. \Cref{sec:Result} provides extensive validation of the replay-based continual learning and demonstrates its effectiveness. \Cref{sec:Discussion} and \Cref{sec:Conclusion} present discussions and concluding remarks.

\section{Preparatory knowledge\label{sec:Preparatory-knowledge}}

We first introduce two mainstream paradigms of neural operators, namely data-driven and physics (PDEs)-driven, and then we introduce continual learning.

\subsection{Neural operator\label{subsec:Neural-operator}}

\begin{figure}
	\begin{centering}
		\includegraphics[scale=0.8]{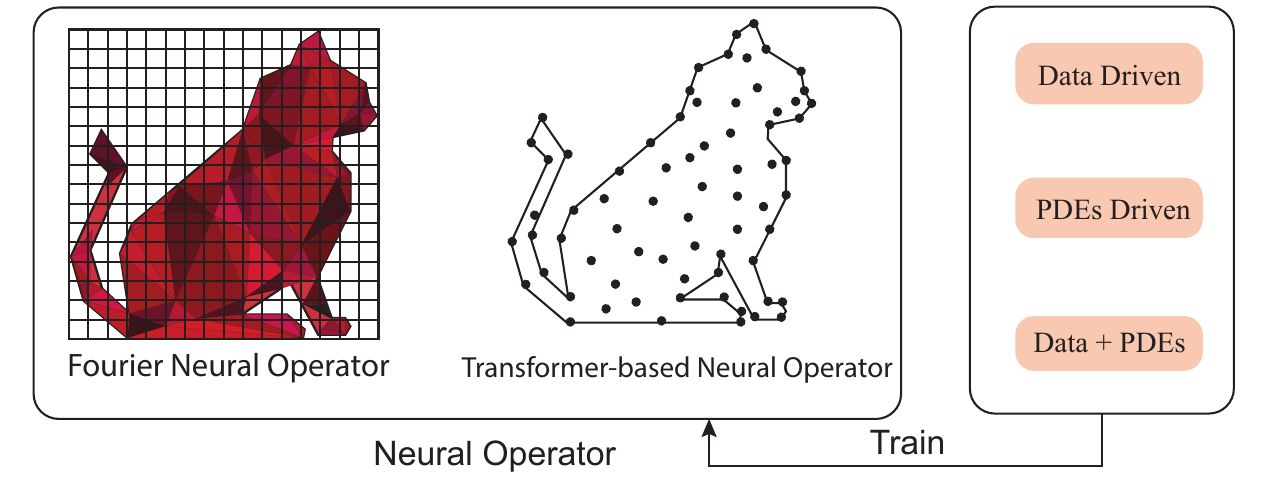}
		\par\end{centering}
	\caption{Introduction to neural operators and several training paradigms\label{fig:Neural_operator_intro}}
\end{figure}

The core of neural operators lies in learning mappings between different function spaces, which has broad applications in science and engineering \citep{kovachki2023neural}. For example, the input function to a neural operator can be the boundary conditions of a partial differential equation, and the output is the function field of interest. Although mathematically functions are infinite-dimensional, in practice we approximate continuous functions by inputting discrete data. Fortunately, neural operators often possess the advantage of discretization invariance. Currently, there are three commonly used types of neural operators: DeepONet \citet{DeepOnet}, FNO \citet{li2020fourier}, and transformer-based neural operators \citet{hao2023gnot,wu2024transolver}. Here we mainly focus on FNO and transformer-based neural operators \citet{hao2023gnot,wu2024transolver}, because empirically DeepONet proposes a generalized architecture, and the original DeepONet faces challenges on some complex problems. In the following we introduce FNO and transformer-based neural operators.

Given that FNO is already a very well-known method, we will not elaborate on it here; for details please refer to the original paper \citet{li2020fourier}. The accuracy and efficiency of FNO are generally quite good, but a key limitation of FNO is the requirement of a regular grid, as shown in \ref{sec:Input-and-output_FNO}. Regarding transformer-based neural operators, we adopt the recently highly influential Transolver \citet{wu2024transolver}. We provide a detailed explanation of Transolver in \ref{sec:transolver_introduction}; details can also be found in the original paper \citet{wu2024transolver}. Training neural operators can generally be divided into three approaches: data-driven, physics (PDEs)-driven, and a hybrid of data and physics. We describe these three training paradigms below.

The most common way to train neural operators is purely data-driven. The data-driven approach is very straightforward. Suppose we have obtained corresponding data $\{\boldsymbol{X}^{(i)},\boldsymbol{Y}^{(i)}\}^{N}_{i=1}$ under different geometries $\boldsymbol{G}$, different materials $\boldsymbol{M}$, and different boundary conditions $\boldsymbol{B}$, where $\boldsymbol{X}^{(i)}$ and $\boldsymbol{Y}^{(i)}$ are the input and output, respectively, and $N$ is the total number of data samples. The optimization process of the purely data-driven approach is:
\begin{equation}
	\begin{aligned}\mathcal{\boldsymbol{H}}_{\boldsymbol{\theta}}(\boldsymbol{x};\boldsymbol{X})= & \arg\min_{\boldsymbol{\theta}}\mathcal{L}_{data}\\
		\mathcal{L}_{data}= & \frac{1}{N}\sum^{N}_{i=1}||\mathcal{\boldsymbol{H}}(\boldsymbol{X}^{(i)};\boldsymbol{\theta})-\boldsymbol{Y}^{(i)}||^{2}
	\end{aligned}
	\label{eq:data_driven}
\end{equation}
where $\boldsymbol{\theta}$ denotes the trainable parameters of $\mathcal{\boldsymbol{H}}(\boldsymbol{X}^{(i)};\boldsymbol{\theta})$, and $||\cdot||^{2}$ is the squared L2 norm. Note that $\mathcal{\boldsymbol{H}}(\boldsymbol{X}^{(i)};\boldsymbol{\theta})$ is the operator approximated by the neural operator, abbreviated as $\mathcal{\boldsymbol{H}}_{\boldsymbol{\theta}}(\boldsymbol{x};\boldsymbol{X}^{(i)})$, and $\boldsymbol{x}$ represents different spatial coordinates. $\boldsymbol{X}^{(i)}$ and $\boldsymbol{Y}^{(i)}$ are functions that contain data at various locations. It is worth noting that the advantage of training operator learning in a data-driven manner is its simplicity and directness. The disadvantage is also obvious: high-quality datasets must be obtained beforehand. In many scenarios data contain errors, the amount of data may be insufficient, and acquiring data can be extremely time-consuming.

Therefore, the traditional data-driven way of training neural operators faces significant challenges, making it highly meaningful to train neural operators using PDEs, i.e., Physics-informed Neural Operator (PINO). Based on the type of neural operator, PINO can be classified into three categories: those with DeepONet architectures \citet{wang2021learning}, those with FNO architectures \citet{li2024physics,eshaghi2025variational}, and those with Transolver architectures \citet{wang2026pfem}. Furthermore, PINO can be divided into two categories according to the form of the PDEs: strong form \citet{wang2021learning,li2024physics} and energy form \citet{eshaghi2025variational,wang2026pfem}. Again, it is emphasized that PINO does not require data to train neural operators; it can train them solely through the governing PDEs. Fortunately, most computational physics problems are governed by corresponding PDEs. The core of computational physics is to solve PDEs numerically. Thus, the core of "PDEs Driven" is to incorporate PDEs into the training of neural operators. The loss function of PINO is typically:
\begin{equation}
	\begin{aligned}\mathcal{\boldsymbol{H}}_{\boldsymbol{\theta}}(\boldsymbol{x};\boldsymbol{X})= & \arg\min_{\boldsymbol{\theta}}\mathcal{L}_{pde}\end{aligned}
	\label{eq:PDEs_driven}
\end{equation}

The PDE forms in PINO are of two types: strong form and energy form. For the strong form:
\begin{equation}
	\mathcal{L}_{pde-s}=\frac{1}{N}\sum^{N}_{i=1}\{\frac{1}{N_{d}}\sum^{N_{d}}_{j=1}||\boldsymbol{P}(\mathcal{\boldsymbol{H}}_{\boldsymbol{\theta}}(\boldsymbol{x}^{(j)};\boldsymbol{X}^{(i)}))||^{2}+\frac{1}{N_{b}}\sum^{N_{b}}_{j=1}||\boldsymbol{I}(\mathcal{\boldsymbol{H}}_{\boldsymbol{\theta}}(\boldsymbol{x}^{(j)};\boldsymbol{X}^{(i)}))||^{2}\}\label{eq:PDEs_driven_strong}
\end{equation}
where $\boldsymbol{P}$ and $\boldsymbol{I}$ represent the governing PDE in the interior domain and the boundary conditions, respectively. $N_{d}$ and $N_{b}$ are the total number of interior points and boundary points, respectively, and $N$ is the total number of data samples. For simplicity, we consider a simple steady-state heat conduction problem with a heterogeneous thermal conductivity field in two dimensions for illustration:
\begin{equation}
	\begin{cases}
		-\nabla\cdot[k(\boldsymbol{x})\nabla T(\boldsymbol{x})]=f(\boldsymbol{x}) & \boldsymbol{x}\in\Omega\\
		T(\boldsymbol{x})=\bar{T}(\boldsymbol{x}) & \boldsymbol{x}\in\Gamma^{T}\\
		k(\boldsymbol{x})\frac{\partial T(\boldsymbol{x})}{\partial\boldsymbol{n}}=\bar{q}(\boldsymbol{x}) & \boldsymbol{x}\in\Gamma^{q}
	\end{cases}.\label{eq:poisson_equation}
\end{equation}
where $k(\boldsymbol{x})$ is the heterogeneous thermal conductivity field; $T(\boldsymbol{x})$ is the temperature field; $f(\boldsymbol{x})$ is the heat source field; $\bar{T}(\boldsymbol{x})$ and $\bar{q}(\boldsymbol{x})$ are the prescribed temperature on the Dirichlet boundary and the prescribed heat flux on the Neumann boundary, respectively. Here the neural operator learns the mapping $\{k(\boldsymbol{x}),f(\boldsymbol{x}),\bar{T}(\boldsymbol{x}),\bar{q}(\boldsymbol{x})\}\overset{\mathcal{\boldsymbol{H}}_{\boldsymbol{\theta}}}{\rightarrow}T(\boldsymbol{x})$. The expression of \Cref{eq:poisson_equation} in the form of \Cref{eq:PDEs_driven_strong} is:
\begin{equation}
	\begin{aligned}\mathcal{L}_{pde-s} & =\frac{1}{N}\sum^{N}_{i=1}\{\frac{1}{N_{d}}\sum^{N_{d}}_{j=1}||\nabla\cdot[k^{(i)}(\boldsymbol{x}^{(j)})\nabla\mathcal{\mathcal{H}}_{\boldsymbol{\theta}}(\boldsymbol{x}^{(j)};k^{(i)},f^{(i)},\bar{T}^{(i)},\bar{q}^{(i)})]+f^{(i)}(\boldsymbol{x}^{(j)})||^{2}\\
		& +\frac{1}{N_{b}}\sum^{N_{bT}}_{j=1}||\mathcal{\mathcal{H}}_{\boldsymbol{\theta}}(\boldsymbol{x}^{(j)};k^{(i)},f^{(i)},\bar{T}^{(i)},\bar{q}^{(i)})-\bar{T}^{(i)}(\boldsymbol{x}^{(j)})||^{2}\\
		& +\frac{1}{N_{b}}\sum^{N_{bq}}_{j=1}||k^{(i)}(\boldsymbol{x}^{(j)})\boldsymbol{n}\cdot\nabla\mathcal{\mathcal{H}}_{\boldsymbol{\theta}}(\boldsymbol{x}^{(j)};k^{(i)},f^{(i)},\bar{T}^{(i)},\bar{q}^{(i)})-\bar{q}^{(i)}(\boldsymbol{x}^{(j)})||^{2}\}
	\end{aligned}
	\label{eq:strong_form_PDEs}
\end{equation}
where $N_{bT}$ and $N_{bq}$ are the total numbers of points on the Dirichlet boundary and Neumann boundary, respectively.

We can also use the variational principle to write the loss function in energy form:
\begin{equation}
	\begin{aligned}\mathcal{L}_{pde-v} & =\frac{1}{N}\sum^{N}_{i=1}\{\int_{\Omega}\frac{1}{2}k^{(i)}[\nabla\mathcal{\mathcal{H}}_{\boldsymbol{\theta}}(\boldsymbol{x};k^{(i)},f^{(i)},\bar{T}^{(i)},\bar{q}^{(i)})]\cdot[\nabla\mathcal{\mathcal{H}}_{\boldsymbol{\theta}}(\boldsymbol{x};k^{(i)},f^{(i)},\bar{T}^{(i)},\bar{q}^{(i)})]d\Omega\\
		& -\int_{\Omega}f^{(i)}\mathcal{\mathcal{H}}_{\boldsymbol{\theta}}(\boldsymbol{x};k^{(i)},f^{(i)},\bar{T}^{(i)},\bar{q}^{(i)})d\Omega\\
		& -\int_{\Gamma^{q}}\bar{q}^{(i)}\mathcal{\mathcal{H}}_{\boldsymbol{\theta}}(\boldsymbol{x};k^{(i)},f^{(i)},\bar{T}^{(i)},\bar{q}^{(i)})d\Gamma\}
	\end{aligned}
	\label{eq:energy_form_PDEs}
\end{equation}

It is worth noting that the choice between the strong form $\mathcal{L}_{pde-s}$ and the energy form $\mathcal{L}_{pde-v}$ depends on the nature of the PDEs. It is worth noting that the applicability of energy-based formulations is not restricted to steady-state or quasi-static problems. For variational transient problems, an incremental energy functional can often be constructed through time discretization, so that the solution at each time step is obtained by minimizing the corresponding incremental potential. Such an energy minimization formulation generally requires appropriate coercivity of the functional, which is satisfied by nonlinear associated plasticity and phase-field models, including $J_2$ plasticity and Allen--Cahn equation. For further discussions on DEM-based variational formulations for PDEs, please refer to Appendix A of \citet{wang2025physics}. Empirically, the energy form achieves better efficiency and accuracy than the strong form, mainly because the order of derivatives in the energy form is lower. Moreover, the number of hyperparameters in the energy form is greatly reduced compared to the strong form, making it much easier to train neural operators via the energy form. However, the disadvantage of the energy form is that it involves integration and that not all problems have a corresponding energy form.

The final approach for training neural operators is to combine data and PDEs, namely "Data + PDEs":
\begin{equation}
	\mathcal{L}=\mathcal{L}_{data}+\lambda\mathcal{L}_{pde}
\end{equation}
where $\lambda$ is an additional hyperparameter.

\subsection{Continual learning \label{subsec:Continual-learning}}

We have introduced neural operators above. Generally speaking, once a neural operator is successfully trained, its efficiency during the inference stage is extremely high, enabling real-time computation. Compared with traditional algorithms, the efficiency can often be improved by several thousand or even tens of thousands of times \citet{bi2023accurate}. However, neural operators also face the OOD problem, i.e., when the training and testing sets differ too much, the accuracy drops significantly. We elaborate on the challenges of neural operators in handling OOD problems in \Cref{subsec:NO_OOD_challenge}. Fortunately, continual learning can address the OOD issue, making the study of continual learning very important.

\begin{figure}
	\begin{centering}
		\includegraphics[scale=0.55]{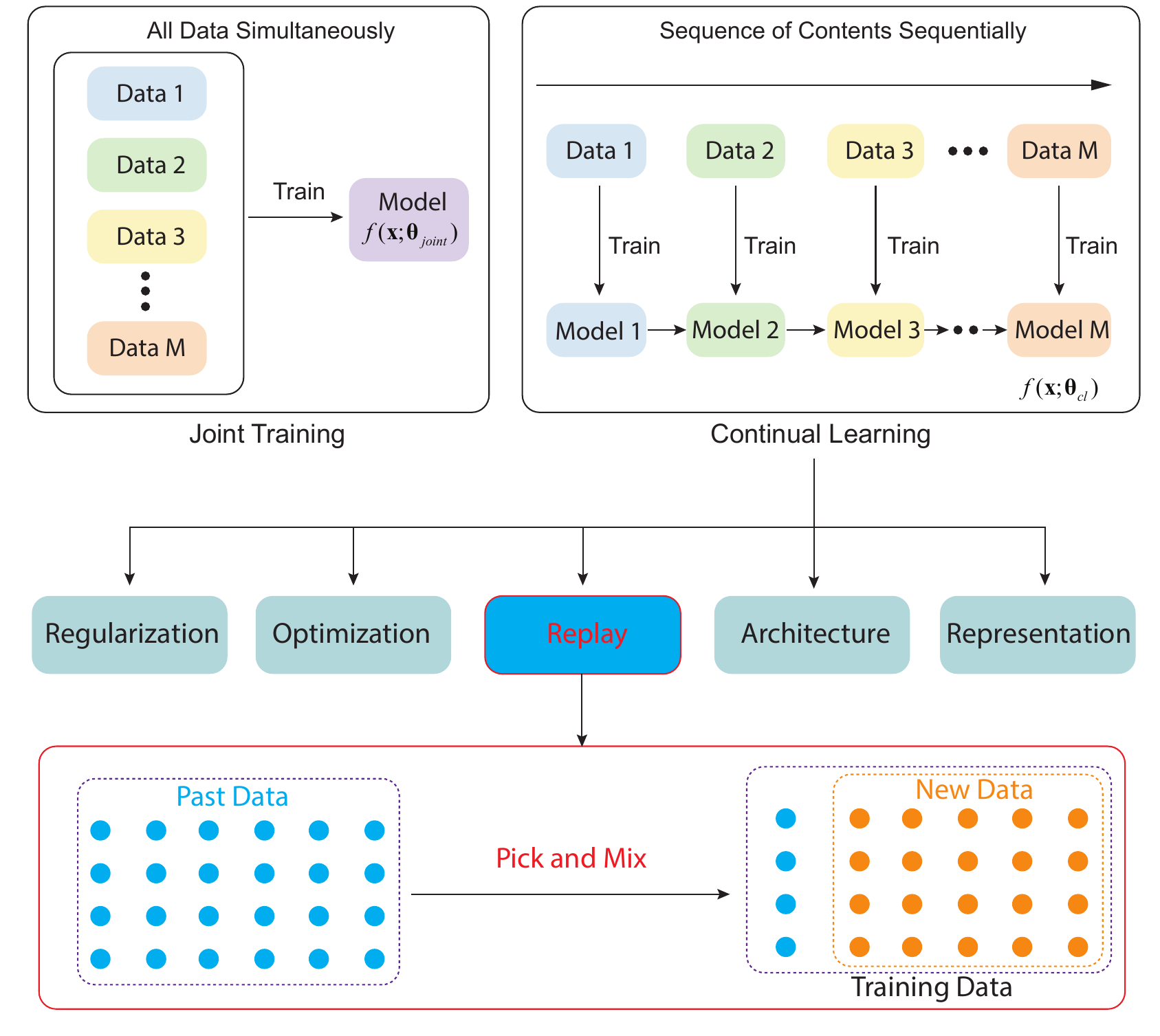}
		\par\end{centering}
	\caption{Comparison between continual learning and joint training: $f(\boldsymbol{x};\boldsymbol{\theta}_{cl})$ and $f(\boldsymbol{x};\boldsymbol{\theta}_{joint})$ are the outputs of the continual learning and joint training models, respectively. Continual learning learns a sequence of contents one by one and aims to approximate the effect of joint training on all tasks simultaneously. Joint learning refers to training on all data at once. Continual learning includes five common categories: regularization-based, replay-based, optimization-based, architecture-based, and representation-based methods. Considering that replay-based methods are simple and highly effective and have been widely used in large language models \citet{abbes2025revisiting}, this work focuses on replay-based continual learning.\label{fig:Continual_learning_and_joint_training}}
\end{figure}

Continual learning aims to enable a model to maintain performance on past tasks as much as possible while adapting to new tasks during the process of receiving new tasks. The ideal outcome of continual learning approaches the effect of joint training on both past and new tasks, as shown in \Cref{fig:Continual_learning_and_joint_training}:
\begin{equation}
	f(\boldsymbol{x};\boldsymbol{\theta}_{cl})\approx f(\boldsymbol{x};\boldsymbol{\theta}_{joint})
\end{equation}
where $\boldsymbol{x}$ is the input, $f$ is the prediction of the model, and $\boldsymbol{\theta}_{cl}$ and $\boldsymbol{\theta}_{joint}$ are the model parameters of continual learning and joint training, respectively. It is worth noting that typically $f(\boldsymbol{x};\boldsymbol{\theta}_{cl})$ cannot achieve the performance of $f(\boldsymbol{x};\boldsymbol{\theta}_{joint})$. The closer $f(\boldsymbol{x};\boldsymbol{\theta}_{cl})$ approximates $f(\boldsymbol{x};\boldsymbol{\theta}_{joint})$, the better the continual learning effect usually is. Joint learning is computationally very expensive, as it requires using all past and new data to train the neural operator. In contrast, continual learning only needs to focus on a small amount of data, thereby significantly reducing both memory usage and computational cost.

Traditional machine learning learns from static data, while continual learning learns from dynamic data. An important issue brought by dynamic data is catastrophic forgetting, which manifests as performance degradation on past tasks while adapting to new tasks. Essentially, continual learning involves a trade-off between learning plasticity and memory stability \citet{wang2024comprehensive}. Learning plasticity refers to the model's ability to generalize to new tasks, whereas memory stability refers to the model's stability on past tasks. Often these two cannot be achieved simultaneously, reminiscent of the no-free-lunch theorem. Continual learning is essentially about finding an optimal balance between learning plasticity and memory stability, so that the model possesses strong generalization ability.

Theoretically, when new data arrives, training on both new and past data together does not lead to catastrophic forgetting. However, in practice, this is extremely costly in terms of computational resources and storage under large-scale data. Therefore, the core of continual learning is to improve the efficiency of model updates when encountering new data. The ideal scenario is to achieve the effect of training on all new and past data by training solely on new data. An excellent continual learning algorithm aims to minimize the use of past data for training and only train on new data. \Cref{fig:Continual_learning_and_joint_training} illustrates the five common categories of continual learning; for details please refer to \citet{wang2024comprehensive}. Empirically, replay-based methods are usually the simplest and yield good results \citet{abbes2025revisiting}. Considering that research on continual learning for neural operators is still limited, we adopt a simple past data, mixes it with new data, and then retrains.

\section{Method\label{sec:Method}}

Continual learning in operator learning requires not only strong adaptability on new data, i.e., possessing learning plasticity, but also the avoidance of catastrophic forgetting on past data, i.e., having memory stability.

In this section, we first propose several categories of problems that continual learning needs to address in neural operators, and then we present a replay-based continual learning approach for neural operators.

\subsection{Classification of continual learning problems in neural operators}

\begin{figure}
	\begin{centering}
		\includegraphics[scale=0.35]{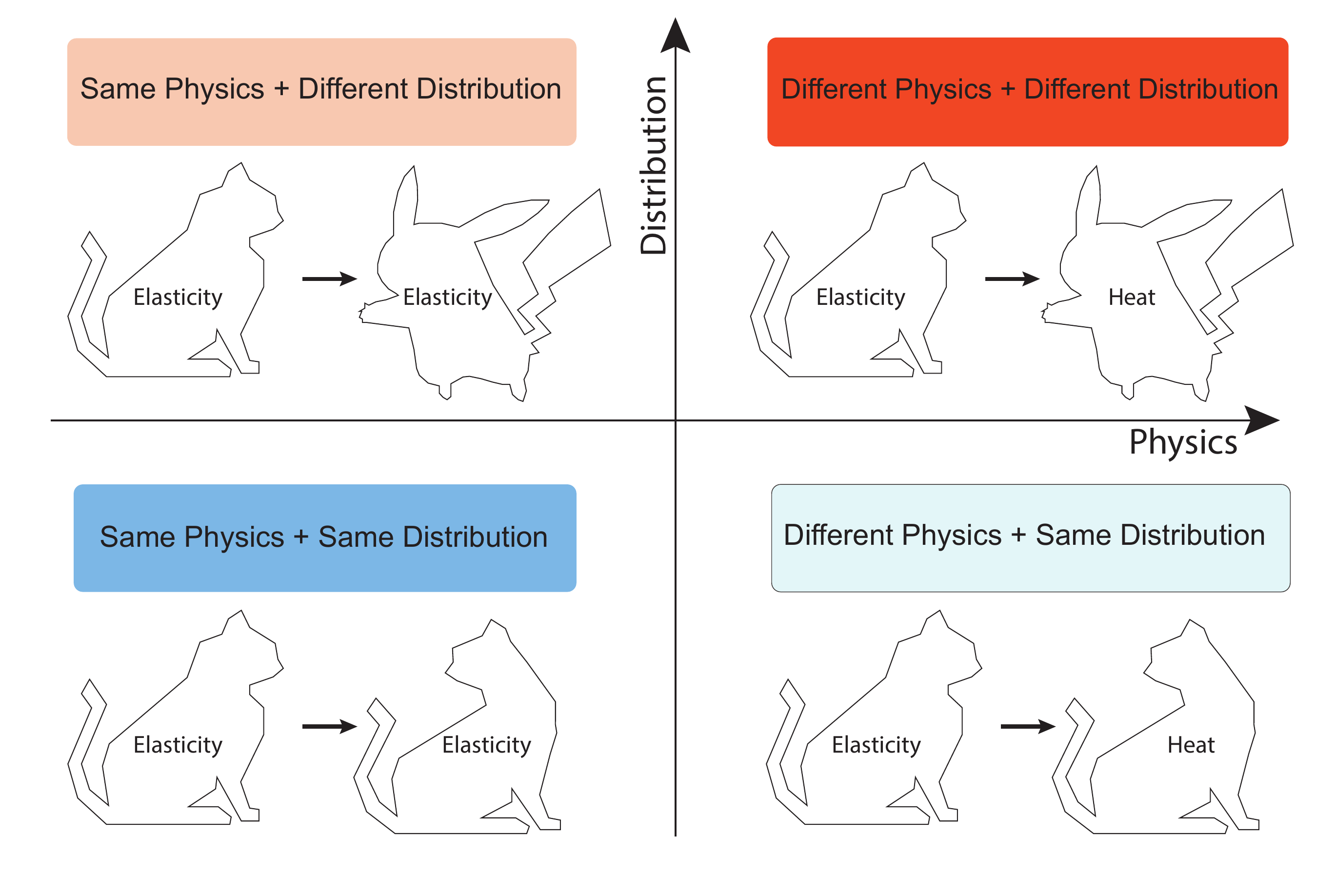}
		\par\end{centering}
	\caption{Illustration of the four types of problems that continual learning needs to handle in operator learning, categorized by differences in physical equations and input distributions. The greater the difference, the more necessary continual learning becomes.\label{fig:classification_continual_learning_NOs}}
\end{figure}

In operator learning, we classify the problems that continual learning needs to handle into the following four types, as illustrated in \Cref{fig:classification_continual_learning_NOs}. For simplicity, we denote past data as $D_{past}=\{X_{past},Y_{past}\}$ and new data as $D_{new}=\{X_{new},Y_{new}\}$. $X$ represents the input of operator learning, such as geometry, material, and boundary conditions; $Y$ is the output of operator learning, such as the displacement field in elasticity.
\begin{itemize}
	\item Same physics and same distribution: The new data and past data share the same physical equations, and the distributions are completely identical, i.e., $p(D_{past})=p(D_{new})$. For instance, $X_{past}$ and $X_{new}$ are samples generated under the same probability density function. If $D_{past}$ is sufficiently large, the neural operator usually performs well on $D_{new}$ without special training.
	\item Same physics and different distribution: The physical equations are the same, but the distributions of new and past data differ, i.e., $p(D_{past})\neq p(D_{new})$. For example, $X_{past}$ and $X_{new}$ are samples generated under different probability density functions. Typically, the larger the gap between $p(D_{past})$ and $p(D_{new})$, the worse the neural operator performs on $D_{new}$.
	\item Different physics and same distribution: The physical equations differ, but the distribution of input $X$ is the same, i.e., $p(X_{past})=p(X_{new})$, while $p(Y_{past})\neq p(Y_{new})$. For example, solving different physical equations on the same structure. Usually, the neural operator needs to take additional information about the different equations as input \citep{zhou2024unisolver}.
	\item Different physics and different distribution: The physical equations differ, and the distribution of input $X$ also differs, i.e., $p(X_{past})\neq p(X_{new})$ and $p(Y_{past})\neq p(Y_{new})$. This is typically the most challenging case. There have been some attempts \citep{zhou2024unisolver,yang2023context}, but challenges still remain.
\end{itemize}
It is worth noting that the above classification is based on data-driven operator learning. For physics-informed learning, labels $Y$ are not required. Generally speaking, regardless of whether it is data-driven or PDE-driven operator learning, performance on new data $D_{new}$ deteriorates in the latter three scenarios. Therefore, continual learning is crucial.

\subsection{A Replay-Based Continual Learning in Neural Operators\label{subsec:replay_LoRA}}

\begin{figure}
	\begin{centering}
		\includegraphics[scale=0.60]{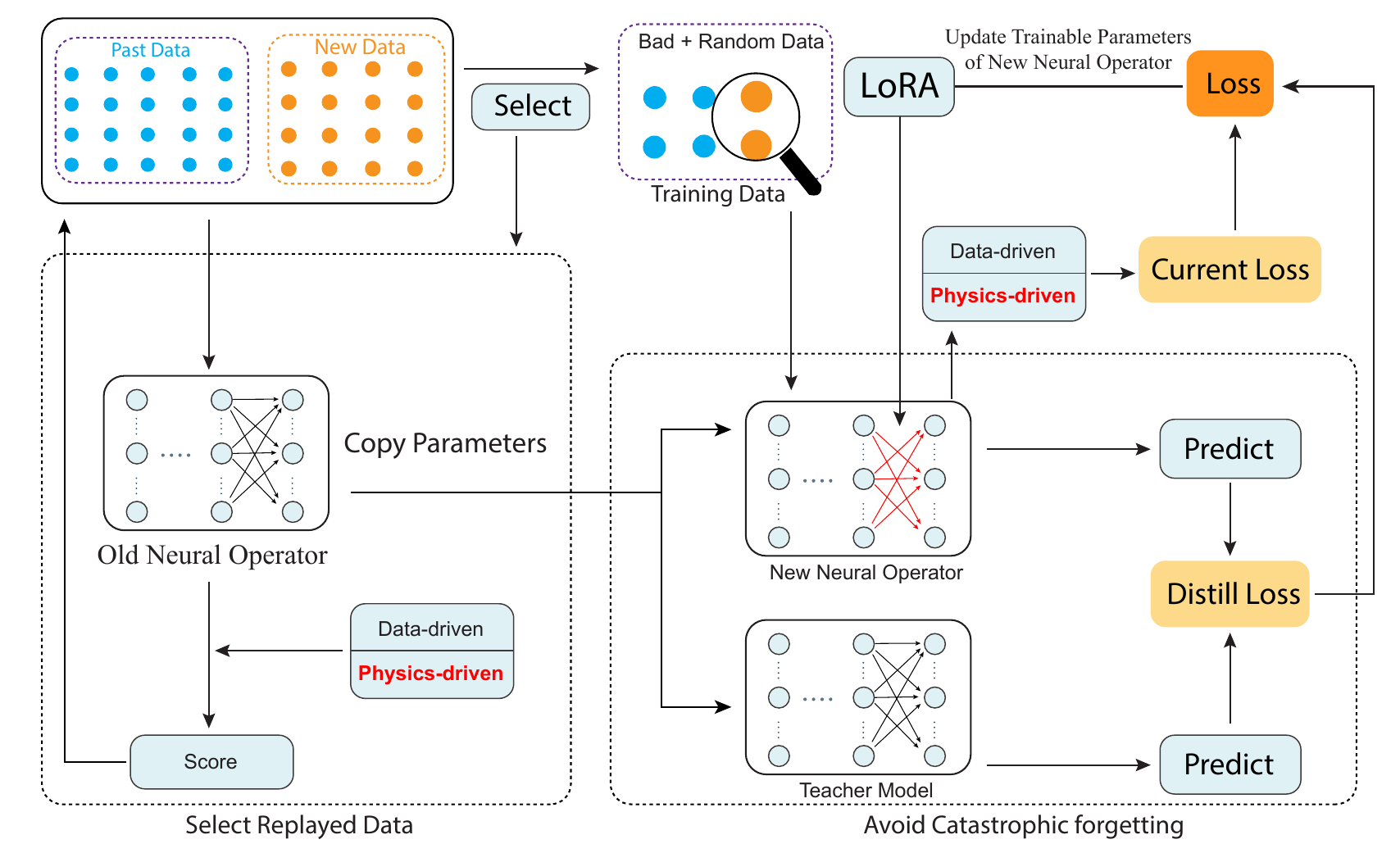}
		\par\end{centering}
	\caption{Illustration of a replay-based continual learning in neural operators: first, mix past and new data to form mixed data. Use the old neural operator model to predict on the mixed data, rank by error to obtain a score. Select data based on the score as the training set, which is often much smaller than the full dataset. Then retrain the neural operator using this training set.\label{fig:Continual-learning_replayed}}
\end{figure}

Given the numerous continual learning methods, we adopt the simplest and most efficient replay-based approach. The core idea is very straightforward: mix the poorly performing data with the new data and retrain, as shown in \Cref{fig:Continual-learning_replayed}. Analogous to human vocabulary learning, before memorizing new words, one mixes previously unmemorized words with new words to maximize time and memory efficiency. Below we elaborate on the method illustrated in \Cref{fig:Continual-learning_replayed}.

Suppose the model has already been trained on past data $D_{past}=\{X_{past},Y_{past}\}$ with parameters $\boldsymbol{\theta}_{past}$. When new data $D_{new}=\{X_{new},Y_{new}\}$ arrive, we first use the old model to predict on both the new data and past data:
\begin{equation}
	\tilde{Y}^{(old)}_{past},\tilde{Y}^{(old)}_{new}=f(X_{past},X_{new};\boldsymbol{\theta}_{past})
\end{equation}

If data-driven, compare $\mathcal{L}(\tilde{Y}^{(old)}_{past},Y_{past})$ and $\mathcal{L}(\tilde{Y}^{(old)}_{new},Y_{new})$, where $\mathcal{L}$ is the relative error. If PDE-driven, theoretically there are no labels $Y$. In this case, we need to use the PDEs to judge the error, e.g., by inserting the model prediction $f$ into the PDEs:
\begin{equation}
	\boldsymbol{P}(f)=\boldsymbol{E}_{pdes}\label{eq:PDEs_loss_contin}
\end{equation}
where $\boldsymbol{P}$ denotes the PDE physical equations, usually in strong or energy form. Then we analyze $\boldsymbol{E}_{pdes}$ to assess the error, requiring a normalization of $\boldsymbol{E}_{pdes}$, because different problems have different loads and scales, i.e., we need to establish a relative value through normalization. As an example, we take the Darcy problem shown in \Cref{eq:Darcy}:
\begin{equation}
	E_{err}=\sqrt{\frac{1}{N}\sum^{N}_{i=1}\frac{\{\nabla\cdot[k(x_{i},y_{i})\nabla f(x_{i},y_{i})]+q(x_{i},y_{i})\}^{2}}{q(x_{i},y_{i})^{2}}}\label{eq:strong_pdes_error}
\end{equation}

For the energy form, the energy must be normalized:
\begin{equation}
	E_{err}=\frac{\frac{1}{2}\int_{\Omega}k(x,y)[\nabla f(x,y)]\cdot[\nabla f(x,y)]dV-\int_{\Omega}q(x,y)f(x,y)dV}{A\int_{\Omega}q(x,y)f(x,y)dV},
\end{equation}
where A is the area of the domain.

We rank the errors $E_{err}$ separately for past data and new data, selecting the worst $M$ samples as "bad data", and then randomly select $N$ samples from the remaining data as "random data". The $M$ "bad data" and $N$ "random data" together form the training set $D_{mix}$ for retraining the neural operator. Below we detail the replay-based continual learning procedure for training the neural operator.

Training the neural operator with $D_{mix}$ is highly susceptible to catastrophic forgetting, meaning we need to enhance the memory stability of continual learning. Our approach is to split the neural operator into two models: the "new neural operator" and the "teacher model". Note that both the "new neural operator" and the "teacher model" are initialized with parameters $\boldsymbol{\theta}_{past}$; the difference is that the parameters of the "new neural operator" are trainable, while those of the "teacher model" are frozen and not updated. Use $D_{mix}$ as input to both the "new neural operator" and the "teacher model", and construct the distillation loss:
\begin{equation}
	\mathcal{L}_{distill}=\mathrm{MSE}(\tilde{Y}^{(new)}_{mix},\tilde{Y}^{(teacher)}_{mix})
\end{equation}
where $\tilde{Y}^{(new)}_{mix}$ and $\tilde{Y}^{(teacher)}_{mix}$ are the outputs of the "new neural operator" and the "teacher model" respectively, and MSE is the standard mean squared error.

To enhance the model's adaptability to new data, i.e., learning plasticity, we construct a task loss $\mathcal{L}_{current}$ on $\tilde{Y}^{(new)}_{mix}$ using either the data-driven approach in \Cref{eq:data_driven} or the physics-driven approach in \Cref{eq:PDEs_driven} as introduced in \ref{subsec:Neural-operator}. Finally, $\mathcal{L}_{distill}$ and $\mathcal{L}_{current}$ are combined to form the loss function $\mathcal{L}$:
\begin{equation}
	\mathcal{L}=\mathcal{L}_{current}+\lambda\mathcal{L}_{distill}
\end{equation}
where $\lambda$ is a hyperparameter; we set it to $0.3$. Generally it should be chosen according to the proportion of past and new data; if past data are more abundant, $\lambda$ should be larger.

Finally, we fine-tune the model using $\mathcal{L}$. Considering that future physics foundation models \citep{hao2024dpot,zhou2024unisolver,mccabe2025walrus} are highly likely to emerge, we adopt LoRA (Low-Rank Adaptation of Large Language Models) \citep{hu2021lora}, currently the most commonly used fine-tuning technique for large models. LoRA can drastically reduce the number of trainable parameters and is a generalized form of fine-tuning; for details please refer to \citep{wang2025transfer}. We discuss the technical details of LoRA in \ref{sec:LoRA_introduction}.

\subsection{Supervised Fine-Tuning\label{subsec:Supervised-Fine-Tuning}}

\begin{figure}
	\begin{centering}
		\includegraphics[scale= 0.6]{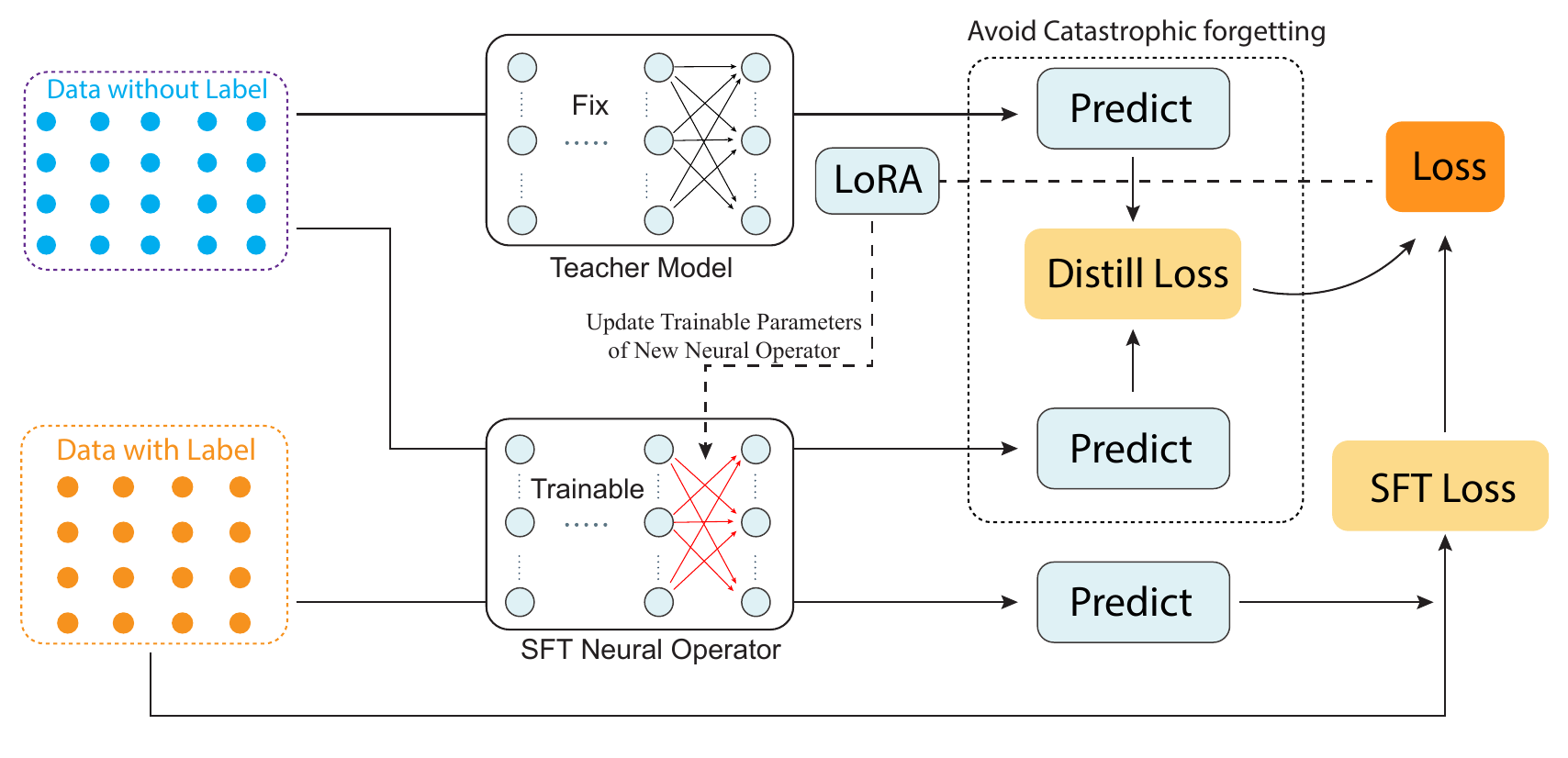}
		\par\end{centering}
	\caption{Illustration of Supervised Fine-Tuning in neural operators: we copy the parameters of the PDE-driven pretrained model to the "teacher model" and the "SFT neural operator." The unlabeled data $D_{left}=\{X_{left}\}$ are fed into both the "teacher model" and the "SFT neural operator" for distillation, constructing the "Distill loss." The labeled high-fidelity data $D_{sft}=\{X_{sft},Y_{sft}\}$ are fed into the "SFT neural operator" to construct the data loss ("SFT loss"). Then LoRA is used to fine-tune the "SFT neural operator."\label{fig:Supervised-Fine-Tuning-operator_learning}}
\end{figure}

As is well known, Supervised Fine-Tuning (SFT) is a key technique for aligning large language models with human instructions \citet{dong2024abilities}, enabling them to better satisfy human preferences. Analogous to large models, we can apply the idea of SFT to neural operators with the goal of further improving predictive accuracy. The core idea is to follow PDE-driven pretraining with data-driven fine-tuning, as illustrated in \Cref{fig:Supervised-Fine-Tuning-operator_learning}. The central concept is to further supervise and fine-tune the model using data, thereby further enhancing its accuracy. Below we describe the specific approach.

Suppose that among the overall data we have a portion of labeled, high-fidelity data $D_{sft}=\{X_{sft},Y_{sft}\}$, and the remaining unlabeled data are $D_{left}=\{X_{left}\}$. The labeled high-fidelity data are used to improve accuracy, while the unlabeled data are used to mitigate catastrophic forgetting. We first copy the parameters of the PDE-driven pretrained model to two identical neural networks: the "teacher model" and the "SFT neural operator." The parameters of the "teacher model" are frozen and not trainable, while the parameters of the "SFT neural operator" are trainable. Then, we feed $D_{left}$ as input to both the "teacher model" and the "SFT neural operator," obtaining the corresponding outputs $\tilde{Y}^{(sft)}_{left}$ and $\tilde{Y}^{(teacher)}_{left}$, and construct a distillation loss $\mathcal{L}_{distill}$:
\begin{equation}
	\mathcal{L}_{distill}=\mathrm{MSE}(\tilde{Y}^{(sft)}_{left},\tilde{Y}^{(teacher)}_{left})
\end{equation}
(Note: there was a typographical error in the original subscript; it should be $\tilde{Y}^{(teacher)}_{left}$ to match the input. The original text had $mix$, but the context indicates $D_{left}$. We'll correct this to maintain consistency.)

Next, we feed the labeled high-fidelity data $D_{sft}=\{X_{sft},Y_{sft}\}$ into the "SFT neural operator" and construct the SFT loss $\mathcal{L}_{sft}$:
\begin{equation}
	\mathcal{L}_{sft}=\mathrm{MSE}(\tilde{Y}^{(sft)}_{sft},Y_{sft})
\end{equation}
Finally, we couple $\mathcal{L}_{distill}$ and $\mathcal{L}_{sft}$ through a hyperparameter:
\begin{equation}
	\mathcal{L}=\mathcal{L}_{sft}+\lambda\mathcal{L}_{distill}
\end{equation}
and then fine-tune the "SFT neural operator" using LoRA. Here $\lambda$ is a hyperparameter, and we set it to $0.3$.

\section{Result\label{sec:Result}}

We test the performance of the replay-based continual learning on three examples: the Darcy flow problem in fluid mechanics, a 2D brain tumor problem in biomechanics, and a 3D TPMS problem in solid mechanics. The Darcy flow example features a clear sequential data stream, where the model receives and learns from group 1 to group 10 in order, simulating a typical continual learning scenario. The 2D brain tumor example is a nonlinear hyperelastic problem with a small amount of new task data, used to examine the method's adaptability under few-shot new data conditions. The 3D TPMS example is a linear elastic problem with a large amount of new task data, used to evaluate the method's stability and forgetting mitigation capability under large-scale new data.

\subsection{Darcy: Challenges of neural operators in handling OOD problems\label{subsec:NO_OOD_challenge}}

Generally speaking, machine learning methods tend to perform poorly on out-of-distribution (OOD) problems. Below we use an example to demonstrate that neural operators also perform poorly on OOD problems.

We first use the Darcy problem to illustrate that neural operators, whether data-driven or PDE-driven, cannot handle OOD problems well.

\begin{figure}
	\begin{centering}
		\includegraphics[scale=0.9]{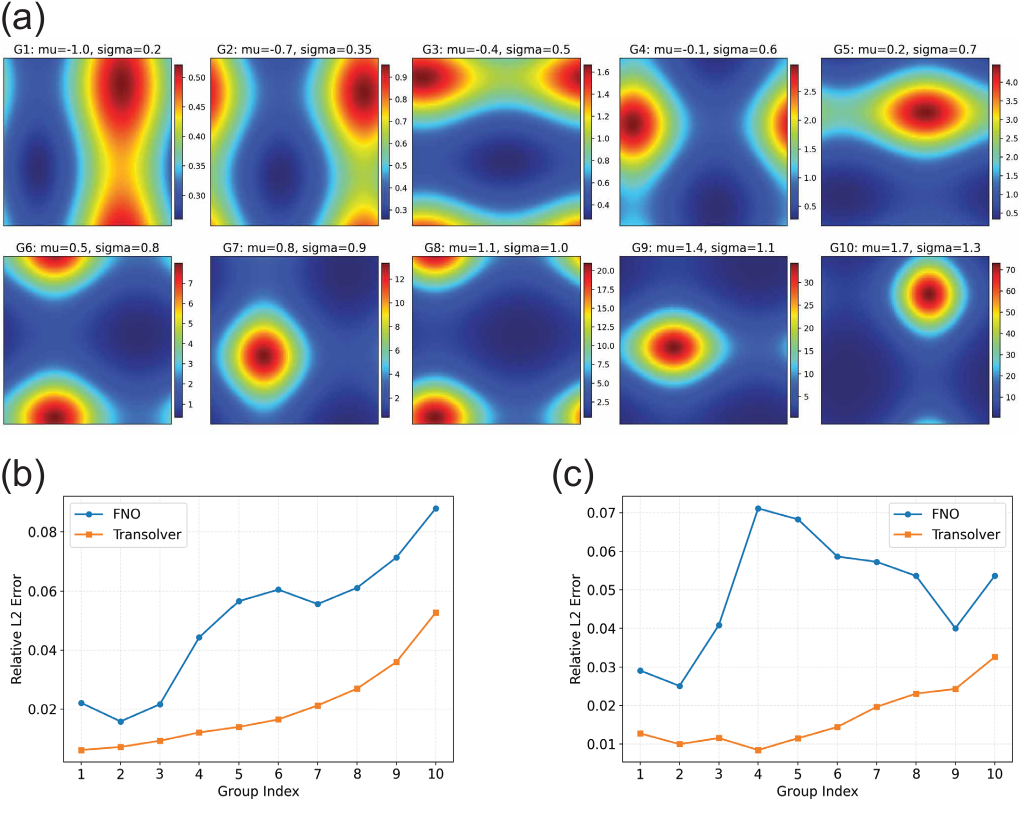}
		\par\end{centering}
	\caption{Joint learning performance on the Darcy problem: (a) Ten groups of permeability field datasets generated from ten different random distributions; (b) Relative errors of data-driven trained FNO and Transolver on each dataset; (c) Relative errors of PDE-driven trained FNO and Transolver on each dataset.\label{fig:Darcy_joint}}
\end{figure}

Darcy flow is a simple classical benchmark in fluid mechanics \citet{li2020fourier}:
\begin{equation}
	\begin{aligned}-\nabla\cdot[k(x,y)\nabla T(x,y)]=q(x,y) & ,\{x,y\}\in[0,1]^{2}\\
		T(x,y)=0 & ,\{x,y\}\in\Gamma
	\end{aligned}
	\label{eq:Darcy}
\end{equation}
where $k$ is the permeability field, $q(x,y)$ is the forcing function, $T$ is the flow velocity, and $\Gamma$ is the boundary. We need the neural operator to learn the mapping $k(\boldsymbol{x})\overset{}{\rightarrow}T(\boldsymbol{x})$. To test the OOD problem, we divide the dataset into 10 groups, each containing 300 samples. We generate the dataset using a Gaussian random field (GRF):
\begin{equation}
	k\sim\exp(\mu+\sigma*GRF)
\end{equation}
where $GRF$ is a standard Gaussian random field. We set different $\{\mu,\sigma\}$ for the 10 groups, specifically $\{\mu,\sigma\}=\{(-1.0,0.2),(-0.7,0.35),(-0.4,0.5),(-0.1,0.6),(0.2,0.7),(0.5,0.8),(0.8,0.9),(1.1,1.0),(1.4,1.1),(1.7,1.3)\}$, as shown in \Cref{fig:Darcy_joint}a.

Next we test the performance of FNO and Transolver on the data from the above ten different distributions, i.e., we use all ten groups together as the training set for joint learning. Note that in FNO, we applied group norm \citet{mccabe2025walrus} because traditional global norm performs poorly on data from multiple distributions. Specifically, we perform group norm after the first fully connected layer in FNO. \Cref{fig:Darcy_joint}b and c respectively show the relative errors of FNO and Transolver under data-driven and PDE-driven training, both using 100 epochs. We can see that basically both FNO and Transolver can generalize well across different data distributions. \Cref{tab:NO_darcy} shows that Transolver has fewer parameters than FNO and better generalization capability across different data distributions, suggesting that transformer-based neural operators have the potential to serve as the backbone of future physics foundation models.

\begin{table}
	\caption{Performance of neural operators FNO and Transolver on the Darcy problem using all data for joint learning. Relative error refers to the average error over the 10 groups of data.\label{tab:NO_darcy}}
	
	\centering{}%
	\begin{tabular}{ccccccc}
		\toprule 
		Training way & Layer & Mode or Slice & Parameters & Learning rate & Relative error & Optimizer\tabularnewline
		\midrule 
		FNO + Data & 4 & 16 & 4735073 & 0.001 & 0.0962 & Adam\tabularnewline
		FNO + PDEs & 4 & 16 & 4735073 & 0.001 & 0.0559 & Adam\tabularnewline
		Transolver + Data & 4 & 32 & 372129 & 0.001 & 0.0212 & Adam\tabularnewline
		Transolver + PDEs & 4 & 32 & 372129 & 0.001 & 0.0157 & Adam\tabularnewline
		\bottomrule
	\end{tabular}
\end{table}

\begin{figure}
	\begin{centering}
		\includegraphics{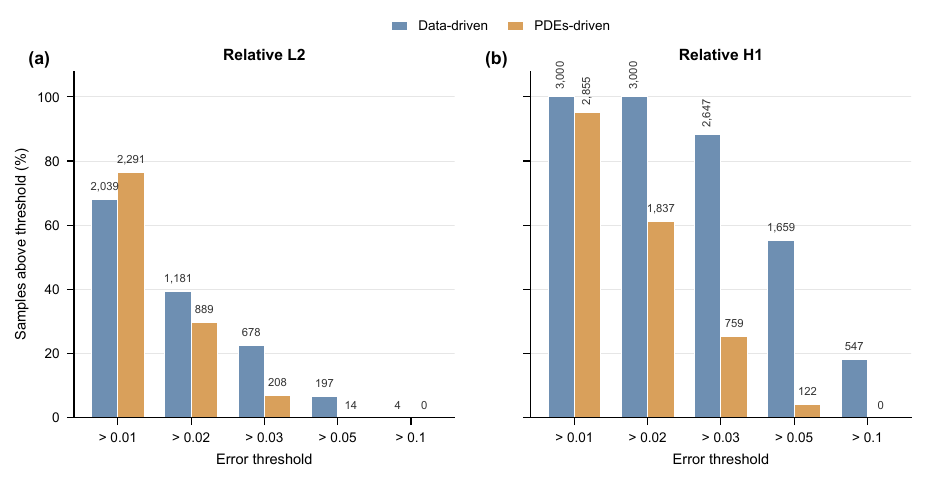}
		\par\end{centering}
	\caption{Comparison between PDEs-driven and data-driven Transolver models for the Darcy problem: (a) relative $\mathcal{L}_{2}$ error; (b) relative $\mathcal{H}_{1}$ error.}
	\label{fig:transolver_PDEs_data}
\end{figure}

\Cref{fig:transolver_PDEs_data} compares the data-driven and PDEs-driven Transolver models across 10 groups with different distributions. We count the number of samples whose relative $\mathcal{L}_{2}$ and $\mathcal{H}_{1}$ errors exceed different thresholds. The results show that the PDEs-driven model significantly reduces the number of high-error samples, with the most pronounced improvement observed in the relative $\mathcal{H}_{1}$ error. This observation is consistent with the optimization objective of the PDEs-driven model, which directly constrains derivative information rather than merely fitting the solution function itself as in the data-driven approach. Most predictions of the PDEs-driven model exhibit errors below $5\%$, demonstrating better stability across different distribution groups.

\begin{figure}
	\begin{centering}
		\includegraphics[scale=0.9]{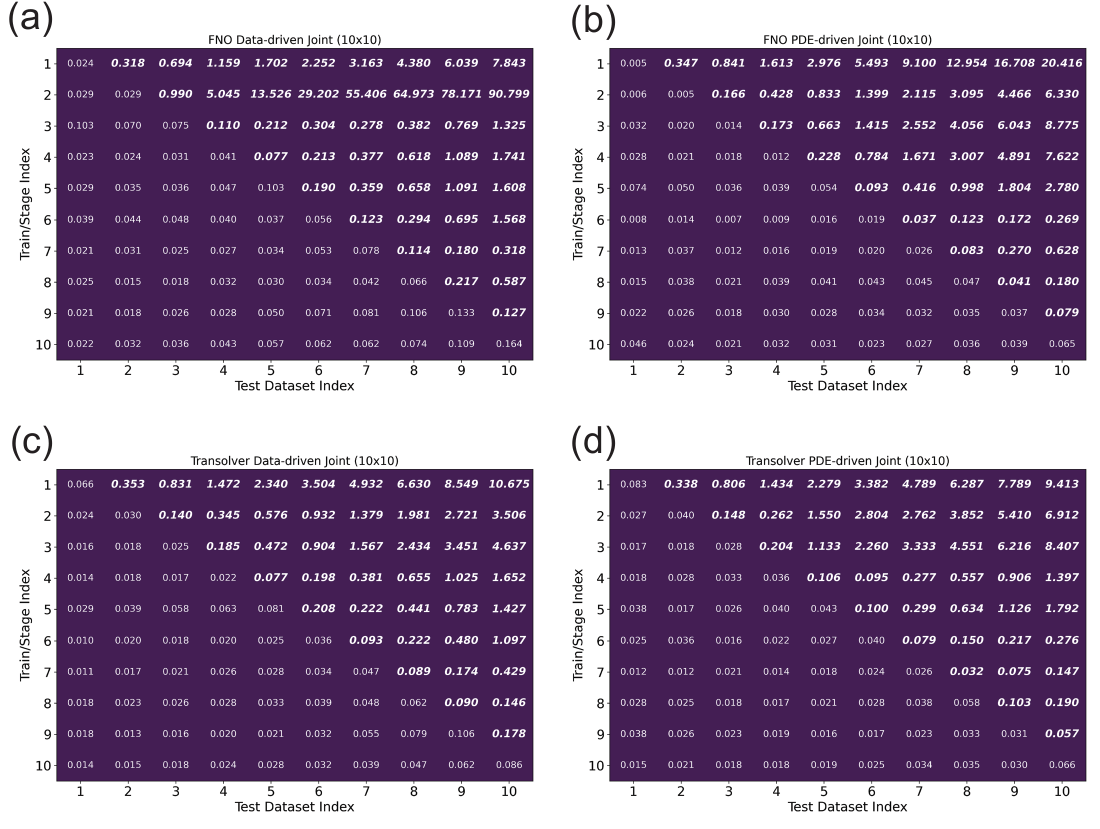}
		\par\end{centering}
	\caption{Performance on in-distribution (ID) and out-of-distribution (OOD) data for the Darcy problem. The numbers indicate relative errors. Bold numbers represent relative errors on OOD data; otherwise, they are on ID data: (a) FNO trained via data-driven; (b) FNO trained via PDE-driven; (c) Transolver trained via data-driven; (d) Transolver trained via PDE-driven.\label{fig:Darcy_joint_OOD}}
\end{figure}

Next we demonstrate the performance of FNO and Transolver on OOD problems. We record the errors of FNO and Transolver after joint learning on $\{D_{1},D_{2},\cdots,D_{m}\}$ and testing on dataset $D_{j}$. If $j\leq m$ it indicates the performance of the neural operator on ID problems, and if $j>m$ it indicates the performance on OOD problems. \Cref{fig:Darcy_joint_OOD} shows that current mainstream neural operators all perform poorly on OOD data. The models were trained for 50 epochs, and the model details are the same as the architectures shown in the joint learning of \Cref{tab:NO_darcy}.

\begin{figure}
	\begin{centering}
		\includegraphics[scale=0.9]{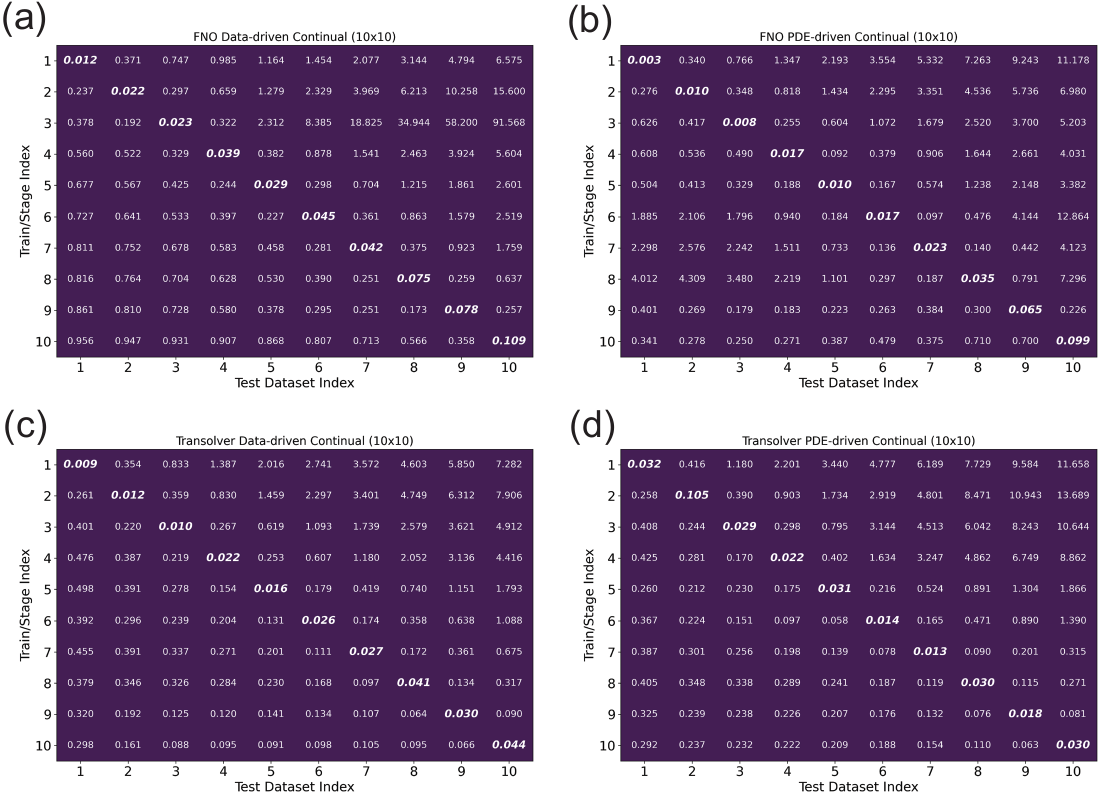}
		\par\end{centering}
	\caption{Performance of fine-tuning on new data for the Darcy problem. Numbers indicate relative errors; bold numbers represent relative errors on new data: (a) FNO trained via data-driven; (b) FNO trained via PDE-driven; (c) Transolver trained via data-driven; (d) Transolver trained via PDE-driven.\label{fig:Darcy_continual_learning}}
\end{figure}

In the real world, we inevitably encounter various OOD data. However, simply fine-tuning on OOD new data usually leads to catastrophic forgetting, as we briefly illustrate below. Let $\{D_{1},D_{2},\cdots,D_{m-1}\}$ be the past data and $D_{m}$ be the new data. We first perform joint learning on $\{D_{1},D_{2},\cdots,D_{m-1}\}$, then fully fine-tune the trained model on $D_{m}$. \Cref{fig:Darcy_continual_learning} shows the effect of fine-tuning only on new data, and we can see that catastrophic forgetting is severe. After fine-tuning on new data, the model's error on past data increases drastically. Therefore, we must mitigate catastrophic forgetting.

The above experiments demonstrate that current neural operators perform poorly on OOD problems, regardless of whether they are data-driven or PDE-driven. Moreover, simply fine-tuning on OOD data causes severe catastrophic forgetting. Therefore, research on continual learning is crucial for handling OOD problems. Considering that PDE-driven neural operator training outperforms data-driven training in both error and efficiency, and that Transolver exhibits stronger generalization capability than FNO, we focus on the replay-based continual learning of the PDE-driven Transolver.

\begin{figure}
	\begin{centering}
		\includegraphics[scale=0.9]{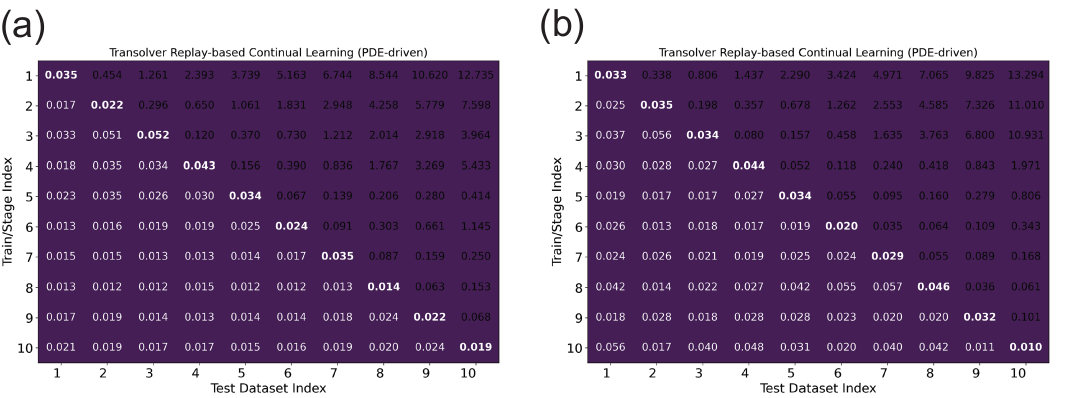}
		\par\end{centering}
	\caption{Performance of replay-based continual learning on Darcy flow. Bold diagonal numbers correspond to relative errors on new data, white numbers correspond to relative errors on training data, and black numbers correspond to data not in the training set: (a) error map of replay-based continual learning, where the model parameters are initialized by joint learning on past data; (b) error map of replay-based continual learning, where the model parameters are initialized by replay-based continual learning on past data. \label{fig:Replay-based_darcy}}
\end{figure}

\Cref{fig:Replay-based_darcy} shows the performance of replay-based continual learning. In \Cref{fig:Replay-based_darcy}a, joint learning is performed on past data $\{D_{1},D_{2},\cdots,D_{m-1}\}$, and then the trained model is subjected to replay-based continual learning on new data $D_{m}$. \Cref{fig:Replay-based_darcy}b shows the effect of sequential learning with replay-based continual learning: the model performs continual learning on past data and then also on new data, which means that new data are continuously handled using replay-based continual learning. We can see that replay-based continual learning performs very well on past data, thus effectively mitigating catastrophic forgetting. At the same time, it exhibits strong adaptability to new data.

\begin{figure}
	\centering
	\includegraphics[scale=0.95]{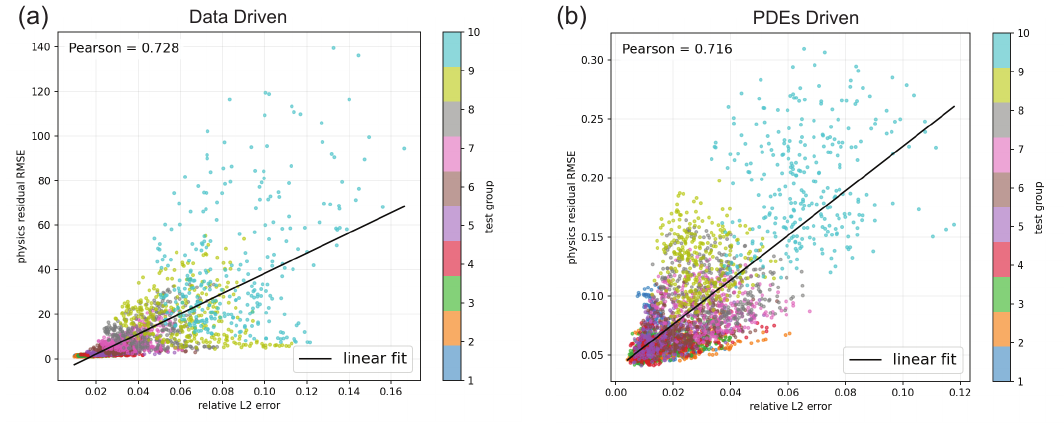}
	\caption{Distribution of the physics-based loss and relative error of the Transolver model on the Darcy flow problem during replay-based continual learning. 
		Points with different colors represent results from different groups. 
		The horizontal axis denotes the relative error, while the vertical axis denotes the physics-based loss.}
	\label{fig:Transolver_data_PDEs_error}
\end{figure}

We further investigate the correlation between the physics-based loss and the relative error. 
Specifically, at a certain training stage of replay-based continual learning, we record the physics-based loss and the corresponding relative error, as shown in \Cref{fig:Transolver_data_PDEs_error}. 
It can be observed that the physics-based loss exhibits a clear correlation with the relative error. 
Therefore, the physics-based loss can be used as an approximate indicator of the relative error, which further provides a practical criterion for selecting samples to be replayed. 
It is worth noting that the physics-based loss is computed according to \Cref{eq:strong_pdes_error}.
In addition, the physics-based loss of the data-driven model shown in \Cref{fig:Transolver_data_PDEs_error}a is relatively large. 
This is because the data-driven model only fits the solution field itself and does not explicitly incorporate derivative information as in the physics-driven training strategy. 
As a result, the $\mathcal{H}^{1}$ error of the data-driven model is much larger than its $\mathcal{L}^{2}$ error.
Interestingly, however, the physics-based loss of the data-driven model still shows a strong correlation with the relative error.

In summary, we demonstrate that our proposed replay-based continual learning has good memory stability as well as good learning plasticity.

\subsection{Brain tumors\label{subsec:brain_tumors}}

\begin{figure}
	\begin{centering}
		\includegraphics[scale=0.9]{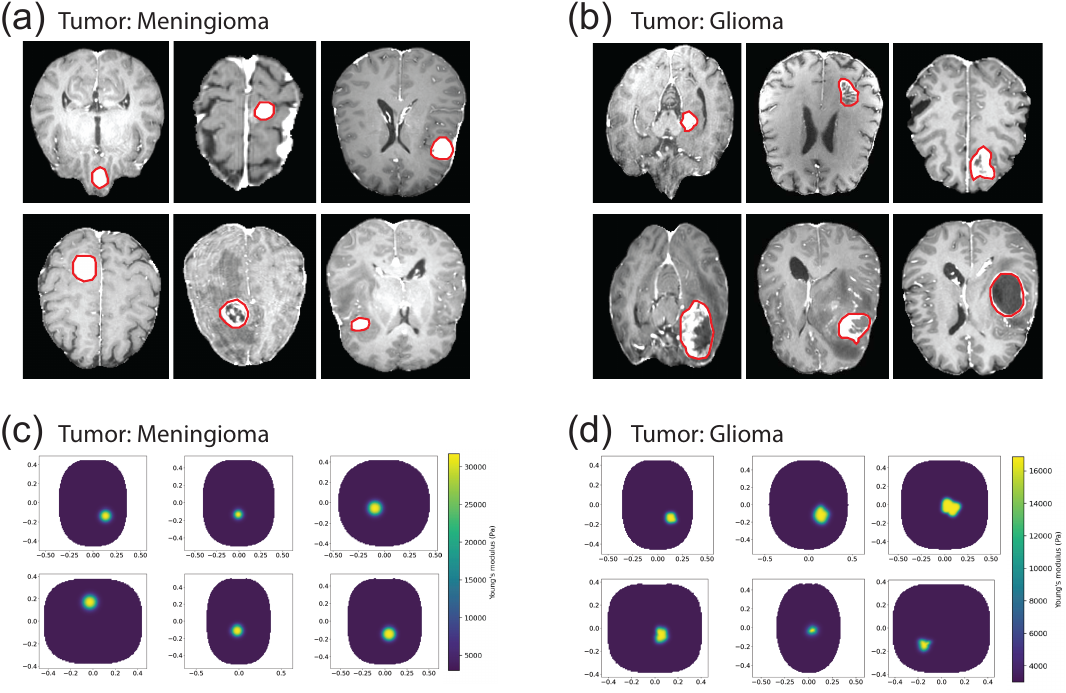}
		\par\end{centering}
	\caption{Shapes of meningiomas and gliomas: (a) Magnetic Resonance Imaging (MRI) of a meningioma, data from \protect\url{https://www.synapse.org/Synapse:syn51514106} \citet{Calabrese_LaBella_2023}; (b) Magnetic Resonance Imaging of a glioma, glioma data from \protect\url{https://www.cancerimagingarchive.net/collection/ucsf-pdgm/} \citet{calabrese2022university}; (c) Distribution of Young's modulus for the meningioma; (d) Distribution of Young's modulus for the glioma.\label{fig:tumor_MRI}}
\end{figure}

Brain tumors represent an important biomedical problem \citet{deangelis2001brain}. We focus on the mechanical behavior of brain tumors, because the sustained growth of a tumor within brain tissue causes deformation of the surrounding tissue and may lead to brain function impairment \citet{ciasca2016nano}. The elastic modulus of the living brain was measured in \citet{chauvet2016vivo}; we do not distinguish between grey and white matter \citet{miller2000mechanical} and uniformly assign an elastic modulus of $E=7.3\,\text{kPa}$. Considering that the elastic moduli of different types of brain tumors vary greatly \citet{bunevicius2020mr}, we consider common meningiomas and gliomas, whose shapes are shown in \Cref{fig:tumor_MRI}a and b.

Hyperviscoelastic, hyperviscoelastic, and linear elasticity material models can all be employed in the brain, yet \citet{wittek2009unimportance} pointed out that different constitutive models have very little effect on practical results, and usually a simple constitutive model is sufficient. Therefore, we adopt the commonly used hyperelastic Neo-Hookean constitutive model to simulate brain tissue \citet{lin2026physics}:
\begin{equation}
	\psi(\boldsymbol{F})=\frac{G}{2}(I_{1}-2-2\ln J)+\frac{\lambda}{2}(\ln J)^{2}\label{eq:biological_energy}
\end{equation}
where $\lambda$ and $G$ are the Lamé parameters
\begin{equation}
	\begin{cases}
		\lambda & =\dfrac{\upsilon E}{(1+\upsilon)(1-2\upsilon)}\\[6pt]
		G & =\dfrac{E}{2(1+\upsilon)}
	\end{cases},
\end{equation}
with $E$ and $\upsilon$ being the Young's modulus and Poisson's ratio, respectively. Furthermore, $\boldsymbol{F}=\partial\boldsymbol{x}/\partial\boldsymbol{X}$ is the deformation gradient, $\boldsymbol{x}$ is the spatial coordinate, $\boldsymbol{X}$ is the material coordinate, and the relationship between spatial and material coordinates is $\boldsymbol{x}=\boldsymbol{X}+\boldsymbol{u}$, where $\boldsymbol{u}$ is the displacement field. $I_{1}=\mathrm{trace}(\boldsymbol{C})$, $J=\mathrm{det}(\boldsymbol{C})$, and $\boldsymbol{C}=\boldsymbol{F}^{T}\boldsymbol{F}$ is the Green deformation tensor. Generally, meningiomas are of regular, round shape, and we set their Young's modulus to $E=33.1\,\text{kPa}$; gliomas are typically geometrically irregular, and we set their Young's modulus to $E=17.5\,\text{kPa}$. These elastic modulus settings are based on \citet{chauvet2016vivo}. Considering that most biological materials are nearly incompressible, we set the Poisson's ratio to $\upsilon=0.49$ \citet{wittek2009unimportance}. The distributions of the elastic modulus fields of meningiomas and gliomas are shown in \Cref{fig:tumor_MRI}c and d.

In addition, a swelling zone, i.e., edema, forms around the brain tumor. Therefore, we set the growth ratio in the tumor region to simulate the compression of the swelling zone on the normal brain. Specifically, we employ the growth tensor $\boldsymbol{F}_{g}$ from biomechanics \citet{lin2026physics}:
\begin{equation}
	\begin{aligned}\boldsymbol{F} & =\boldsymbol{F}_{e}\boldsymbol{F}_{g}=\frac{\partial\boldsymbol{x}}{\partial\boldsymbol{X}}\\
		\boldsymbol{F}_{g} & =g\boldsymbol{I}
	\end{aligned}
\end{equation}
where $g$ in the growth tensor $\boldsymbol{F}_{g}$ is the growth ratio. For normal tissue, $g=1$; for the tumor, $g>1$, and we set it around $1.1$. Substituting $\boldsymbol{F}_{e}=\boldsymbol{F}\boldsymbol{F}^{-1}_{g}$ into \Cref{eq:biological_energy}, i.e., using $\boldsymbol{F}_{e}$ to compute the hyperelastic strain energy, takes into account the growth effect. Moreover, because the brain is constrained by the skull, we set the entire peripheral boundary $\Gamma$ of the brain as fixed:
\begin{equation}
	\begin{aligned}\mathcal{L} & =\int_{\Omega}\psi(\boldsymbol{F}_{e})dV\\
		\boldsymbol{u}(\boldsymbol{x}) & =0,\;\boldsymbol{x}\in\Gamma
	\end{aligned}
\end{equation}
Due to the displacement constraints at the brain boundary, stress builds up inside the skull when the tumor region grows. We generate simulation data based on the actual shapes of the brain and the tumor, as shown in \Cref{fig:tumor_MRI}c and d; more samples are provided in \ref{sec:sample_tumors}.

Since the brain geometry is not regular, if we were to use FNO, we would need a regular grid, which is explained in detail in \ref{sec:Input-and-output_FNO}. The workaround for FNO is to set values to zero where there is no material, which is clearly very inconvenient, as it is equivalent to finding a larger rectangle to enclose the irregular geometry. Therefore, we use Transolver to handle this problem; Transolver only requires point clouds, making it very suitable for problems with complex geometries \citet{wang2026pfem}. Next, we test the performance of Transolver trained in data-driven and PDE-driven fashions, respectively.

\begin{table}
	\caption{Performance of Transolver on the brain tumor problem. Layer and slice are the architecture details of the Transolver model. Time refers to the average prediction time of Transolver and FEM on the corresponding data. $\mathcal{L}_{dis}$ and $\mathcal{L}_{von}$ are the relative errors of displacement and Von Mises stress, with FEM as the reference solution. The FEM results have been verified through mesh convergence tests and are reliable. ``M'' and ``G'' denote meningioma and glioma. The relative errors are averaged over the test set. $\rightarrow$ indicates training first on past data (meningioma) and then fine-tuning on new data (glioma). In the $\rightarrow$ experiments, the initial parameters are those obtained after training on past data. ``Data'' and ``PDEs'' represent training the neural operator based on data and physical equations, respectively. ``Joint'' means fully mixing new and past data, while ``Replay'' means selecting a portion from both. ``Replay PDEs with SFT'' indicates performing SFT on the model obtained after replay-based continual learning.\label{tab:transolver_tumor}}
	\begin{adjustbox}{max width=\textwidth}
	\centering{}%
	\begin{tabular}{ccccccccc}
		\toprule 
		Training way & Layer & Slice & Learning rate & Time: s (Transolver/FEM) & $\mathcal{L}_{dis}$(M/G) & $\mathcal{L}_{von}$(M/G) & Time: s/epoch & Optimizer\tabularnewline
		\midrule 
		M (Data) & 5 & 32 & 0.0008 & 0.02/24.24 & 0.033/0.249 & 0.037/0.205 & 17.83 & Adam\tabularnewline
		M (PDEs) & 5 & 32 & 0.0008 & 0.02/24.24 & 0.027/0.665 & 0.022/0.315 & 18.44 & Adam\tabularnewline
		M$\rightarrow$G (Joint PDEs) & 5 & 32 & 0.0008 & 0.02/24.24 & 0.0318/0.0835 & 0.0242/0.0641 & 29.37 & Adam\tabularnewline
		M$\rightarrow$G ( PDEs) & 5 & 32 & 0.0008 & 0.02/24.24 & 0.0518/0.0836 & 0.0346/0.0660 & 6.80 & Adam\tabularnewline
		M$\rightarrow$G ( PDEs with SFT) & 5 & 32 & 0.0008 & 0.02/24.24 & 0.0397/0.0699 & 0.0291/0.0668 & 6.46 & Adam\tabularnewline
		\bottomrule
	\end{tabular}
\end{adjustbox}
\end{table}

We first simulate the shapes of the brain and brain tumors based on real MRI, generating 1000 meningioma and 200 glioma samples, as shown in \Cref{fig:tumor_MRI}c and d, respectively. To test the model's performance on OOD problems, we first train the model on 800 meningioma samples and test on 200 meningioma samples. The details of the model parameters are given in the first and second experiments of \Cref{tab:transolver_tumor}. We train Transolver using data-driven and PDE-driven approaches, which have been introduced in detail in \Cref{subsec:Neural-operator}.

\begin{figure}
	\begin{centering}
		\includegraphics[scale=0.9]{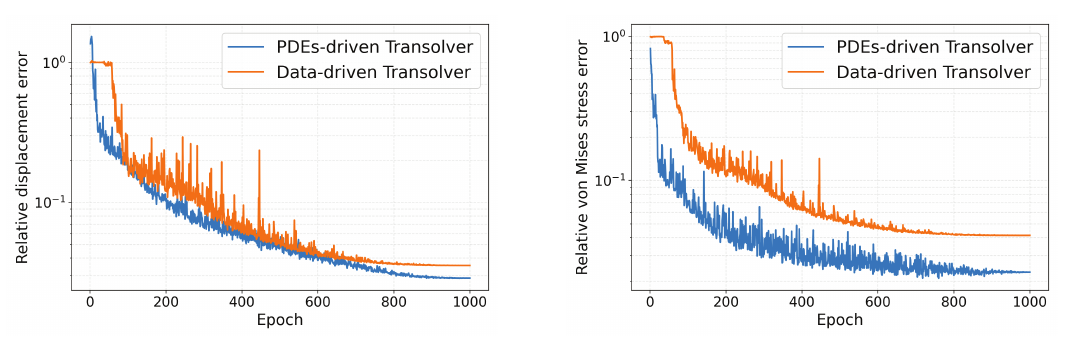}
		\par\end{centering}
	\caption{Evolution trends of relative errors on the test set for the meningioma problem when training the Transolver neural operator with data-driven and PDE-driven approaches. Left: relative error of displacement; Right: relative error of Von Mises stress.\label{fig:Meningioma_relative_error}}
\end{figure}

\Cref{fig:Meningioma_relative_error} shows the evolution trends of the relative errors for data-driven and PDE-driven training. We can see that the PDE-driven approach yields lower relative errors and converges faster. Moreover, data-driven training requires additional time for data generation; we explain the benefit of PDE-driven training over data-driven training in \Cref{subsec:PDEs_benefit}. \Cref{tab:transolver_tumor} also shows that the average computation time for one meningioma sample is $24.24$ seconds, meaning that 1000 meningioma samples require roughly 7 hours. Training Transolver with data-driven or PDE-driven approaches to convergence (500 epochs) takes about 3 hours. In other words, the data generation time exceeds the neural operator training time, and the PDE-driven approach achieves better accuracy. Therefore, for problems with well-defined PDEs, the PDE-driven approach has almost overwhelming advantages over the data-driven approach in terms of both accuracy and efficiency. Note that the key to the efficiency improvement of the PDE-driven approach lies in explicitly constructing derivatives (e.g., using finite element shape functions) rather than employing automatic differentiation; for details please refer to Appendix A of \citet{wang2026pfem}.

\begin{figure}
	\begin{centering}
		\includegraphics[scale=0.9]{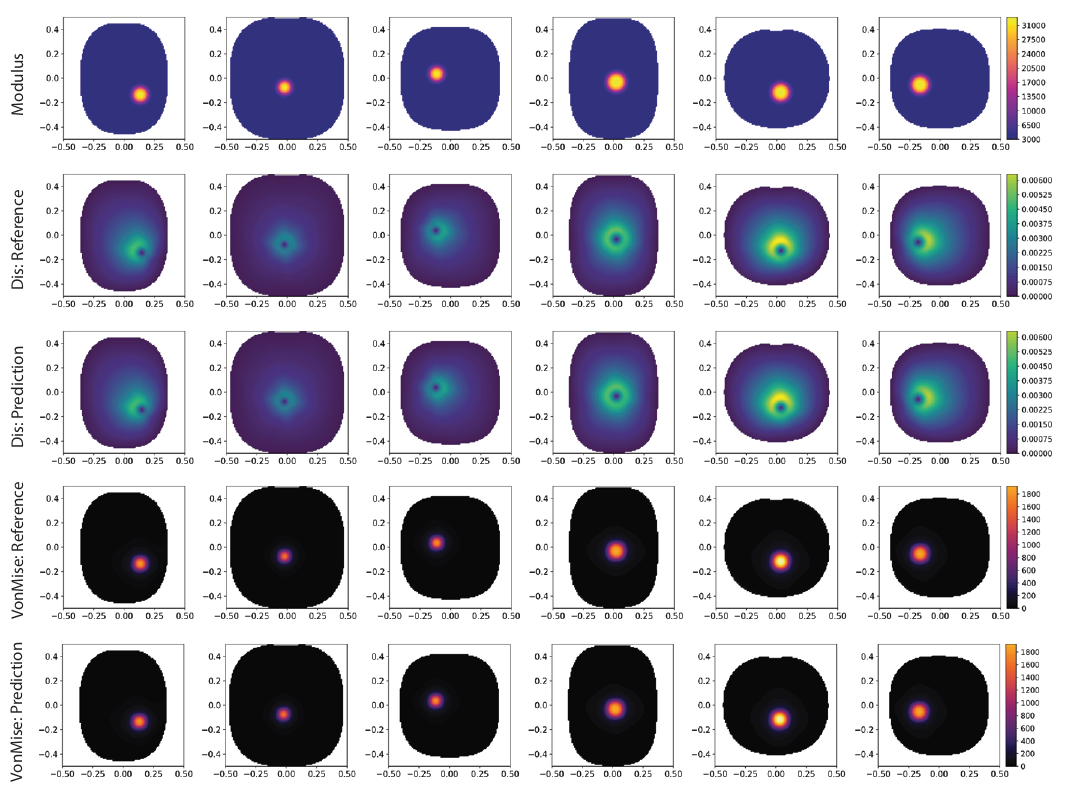}
		\par\end{centering}
	\caption{Prediction results of the PDE-driven Transolver neural operator on the meningioma problem. Rows 1 to 5: distribution of Young's modulus, absolute displacement field from FEM, predicted absolute displacement field from Transolver, Von Mises stress from FEM, and predicted Von Mises stress from Transolver.\label{fig:PDEs_transolver_meningioma}}
\end{figure}

Considering the advantage of the PDE-driven approach over the data-driven approach for the 2D brain tumor problem, we focus only on the PDE-driven approach in the following and no longer consider the data-driven approach. \Cref{fig:PDEs_transolver_meningioma} shows the predicted contours of the displacement and Von Mises stress fields obtained by the PDE-driven Transolver, with FEM as the reference solution. We can see that the PDE-trained Transolver can predict the displacement and stress fields well, and during the inference stage, its efficiency can be improved by thousands of times compared to traditional FEM, as shown in \Cref{tab:transolver_tumor}. This indicates that, after obtaining MRI images, the neural operator can immediately produce the corresponding stress fields. In the future, neural operators are expected to achieve real-time prediction of displacement and stress fields in real MRI scenarios.

Although Transolver performs very well on the meningioma problem, if we switch to the glioma problem (OOD problem), the accuracy drops drastically. For example, in the first and second experiments of \Cref{tab:transolver_tumor}, the first data-driven experiment yields test errors of $0.033$ and $0.249$ on meningioma and glioma, respectively; the second PDE-driven Transolver experiment yields test errors of $0.027$ and $0.665$ on meningioma and glioma, respectively. To improve the accuracy on OOD problems, the simplest method is to mix past and new data for joint learning, as shown in the third experiment of \Cref{tab:transolver_tumor}. Although the accuracy is good, the computational cost is very high, with the iteration time per epoch being around 30 seconds. Therefore, we employ replay-based continual learning to improve training efficiency, as shown in the fourth experiment of \Cref{tab:transolver_tumor}. The iteration time per epoch of the replay-based continual learning is around 6 seconds, and the accuracy can approach that of joint learning. Note that the stopping criterion for both joint and continual learning is when the relative error on past and new data falls below $0.085$. The training parameters for continual learning are set as follows: $10\%$ of the past data are selected, of which $8\%$ are the samples with the highest relative errors and $2\%$ are random samples; $80\%$ of the new data are selected, of which $64\%$ are the samples with the highest relative errors and $16\%$ are random samples. The rank of LoRA is set to 8.

\begin{figure}
	\begin{centering}
		\includegraphics[scale=0.9]{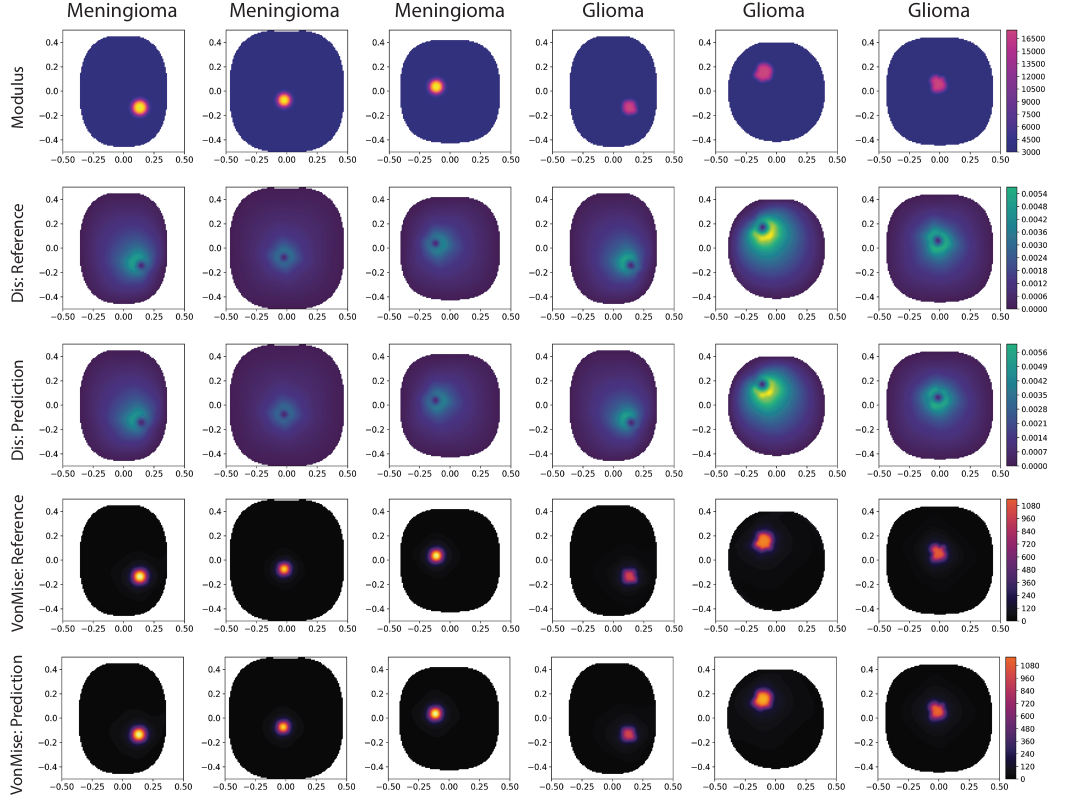}
		\par\end{centering}
	\caption{Prediction results of the PDE-trained Transolver neural operator on meningioma and glioma problems. Rows 1 to 5: distribution of Young's modulus, absolute displacement field from FEM, predicted absolute displacement field from Transolver, Von Mises stress from FEM, and predicted Von Mises stress from Transolver. The first three columns show meningioma results, and the last three columns show glioma results.\label{fig:PDEs_transolver_meningioma_glioma}}
\end{figure}

\begin{figure}
	\begin{centering}
		\includegraphics[scale=0.9]{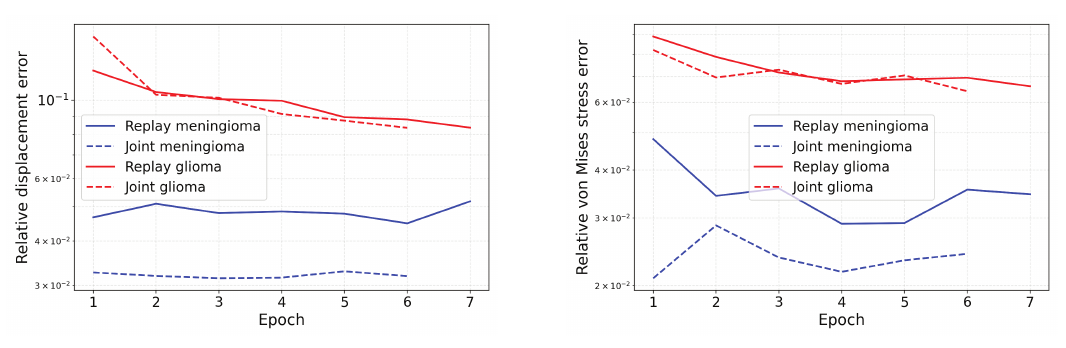}
		\par\end{centering}
	\caption{Evolution trends of relative errors of joint learning and replay-based continual learning on past data (meningioma) and new data (glioma). Left: relative error of displacement; Right: relative error of Von Mises stress.\label{fig:Meningioma_Glioma_relative_error}}
\end{figure}

\Cref{fig:PDEs_transolver_meningioma_glioma} shows the performance of replay-based continual learning on past data (meningioma) and new data (glioma). We can see that after replay-based continual learning, Transolver matches the FEM reference solution very well, with relative errors of only about $5\%$. Importantly, the prediction efficiency during inference is improved by thousands of times, as indicated by the ``Time'' column in \Cref{tab:transolver_tumor}. \Cref{fig:Meningioma_Glioma_relative_error} shows the error evolution trends of joint and continual learning. We can see that after about 7 epochs the model can achieve relative errors below $0.085$ on past and new data. Note that the per-iteration time of replay-based continual learning is significantly lower than that of joint learning, as shown in the fourth experiment of \Cref{tab:transolver_tumor}. The initial parameters of both the replay-based continual learning and joint learning models are the parameters obtained after training on meningioma data.

It is worth noting that because we are PDE-driven, we do not know the labels of the training data in advance. Here we use \Cref{eq:PDEs_driven} for evaluation; specifically, we use the energy functional to predict the error of each sample. Since this problem involves pure displacement loading (fixed boundaries), no body force, and roughly similar overall shape sizes, we directly use the unnormalized energy norm to rank the errors for convenience.

\begin{figure}
	\begin{centering}
		\includegraphics[scale=0.9]{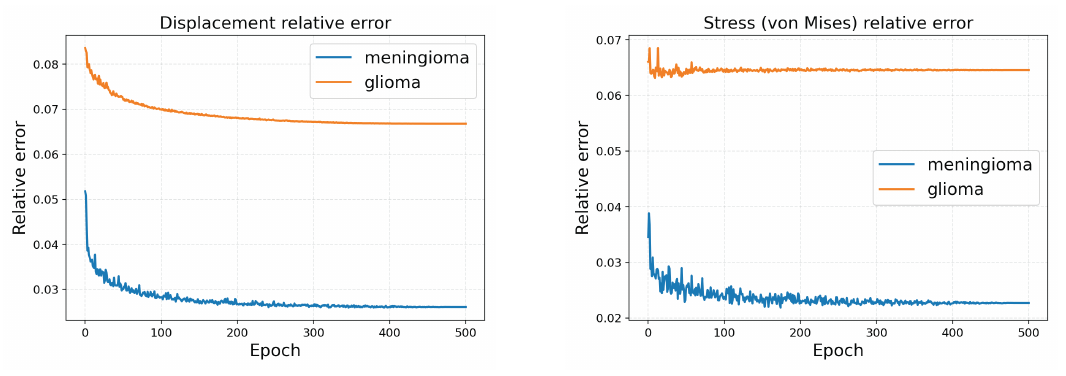}
		\par\end{centering}
	\caption{Effect of Supervised Fine-Tuning (SFT) on the PDE-trained Transolver neural operator. Left: relative error of displacement; Right: relative error of Von Mises stress. We select $10\%$ of the worst-performing samples for SFT fine-tuning, and the remaining $90\%$ of the data are used for model distillation.\label{fig:SFT_tumors}}
\end{figure}

Next, we test the performance of Supervised Fine-Tuning (SFT) as described in \Cref{subsec:Supervised-Fine-Tuning} on the tumors problem. We apply SFT on top of the model obtained after replay-based continual learning. We evaluate the errors on 1200 samples, including 1000 meningioma and 200 glioma samples. We select the $10\%$ of samples with the largest errors, then fine-tune the model using data-driven learning on these $10\%$ samples, while the remaining $90\%$ of the samples are used for distillation. \Cref{fig:SFT_tumors} shows that the model performance is significantly improved after SFT, and the quantitative results are given in the fifth experiment of \Cref{tab:transolver_tumor}.

\subsection{TPMS}

\begin{table}
	\caption{Performance of Transolver on 3D TPMS. Layer and slice are the architecture details of the Transolver model. Time refers to the average prediction time of Transolver and FEM on all data (``Solid-networks'' + ``Sheet-networks''). $\mathcal{L}_{dis}$ is the relative error of the displacement field (the displacement component along the tensile direction in the x-direction), with FEM as the reference solution. The first experiment uses ``Solid-networks'' as the training set; the second uses ``Sheet-networks''; the third uses both ``Solid-networks'' and ``Sheet-networks''. The fourth and sixth experiments have past data as ``Solid-networks'' and new data as ``Sheet-networks''. The fifth and seventh experiments have past data as ``Sheet-networks'' and new data as ``Solid-networks''. All results above are after convergence.\label{tab:transolver_3D_tpms}}
	\begin{adjustbox}{max width=\textwidth}
	\centering{}%
	\begin{tabular}{cccccccc}
		\toprule 
		Training data & Layer & Slice & Learning rate & Time: s (Transolver/FEM) & $\mathcal{L}_{dis}$(Solid/Sheet) & Time: s/epoch & Optimizer\tabularnewline
		\midrule 
		Solid & 4  & 64  & 0.0005 & 0.074/150.0 & 0.0863/3.809 & 1067 & Adam\tabularnewline
		Sheet & 4  & 64  & 0.0005 & 0.074/150.0 & 1.152/0.0729 & 1111 & Adam\tabularnewline
		Solid+Sheet (joint learning) & 4  & 64  & 0.0005 & 0.074/150.0 & 0.0733/0.0696 & 2261 & Adam\tabularnewline
		Solid$\rightarrow$Sheet (joint learning) & 4  & 64  & 0.0005 & 0.074/150.0 & 0.0741/0.0783 & 2236 & Adam\tabularnewline
		Sheet$\rightarrow$Solid (joint learning) & 4  & 64  & 0.0005 & 0.074/150.0 & 0.0773/0.0667 & 2259 & Adam\tabularnewline
		Solid$\rightarrow$Sheet (continual learning) & 4  & 64  & 0.0005 & 0.074/150.0 & 0.0735/0.0644 & 1683 & Adam\tabularnewline
		Sheet$\rightarrow$Solid (continual learning) & 4  & 64  & 0.0005 & 0.074/150.0 & 0.0960/0.0773 & 1749 & Adam\tabularnewline
		\bottomrule
	\end{tabular}
\end{adjustbox}
\end{table}

Homogenization plays a fundamental role in multiscale modeling in solid mechanics \citet{hashin1983analysis,guedes1990preprocessing,hassani1998review} and is often closely tied to the Representative Volume Element (RVE). Homogenization is the process of obtaining macroscopic effective material parameters using the RVE, making it an important tool for studying multiscale physical phenomena. Traditional numerical homogenization methods, heavily reliant on finite element analysis, demand significant computational resources, especially for complex geometries, materials, and high-resolution problems. Recently, using operator learning to study the mechanical behavior of multiscale phenomena in solid mechanics has attracted considerable attention \citet{harandi2025spifol,wang2026pretraining,wang2026pfem}. Therefore, in this subsection we demonstrate the performance of continual learning on a 3D solid mechanics homogenization problem.

Among the many homogenization theories, we select the classical numerical homogenization approach proposed by Andreassen et al. \citet{andreassen2014determine}, which aims to calculate the effective macroscopic elastic tensor of periodic composite materials. For the specific homogenization theory, refer to \citet{wang2026pretraining}. In \citet{wang2026pretraining}, data-driven training is used for FNO. Here we attempt to train Transolver purely with physics-driven approach.

The loss function is determined by the principle of minimum potential energy in elasticity:

\begin{equation}
	\begin{aligned}\mathcal{L} & =\frac{1}{6}\sum^{6}_{IJ}\Pi^{(IJ)}\\
		\Pi^{(IJ)} & =\int_{\Omega}\frac{1}{2}E_{klpq}\varepsilon_{kl}(\boldsymbol{X}^{(v)(IJ)})\varepsilon_{pq}(\boldsymbol{X}^{(v)(IJ)})d\Omega-\int_{\Omega}E_{klpq}(\varepsilon^{(0)(IJ)}_{kl})d\Omega
	\end{aligned}
	\label{eq:homo_loss}
\end{equation}
where $\boldsymbol{X}^{(v)}$ is the fluctuation displacement, i.e., our output target. $\boldsymbol{\varepsilon}^{(0)}$ represents the six loading cases applied during homogenization, i.e., 6 unit strain fields. For a detailed introduction of the homogenization theory, please see \ref{sec:homo}.

\begin{figure}
	\begin{centering}
		\includegraphics[scale=0.55]{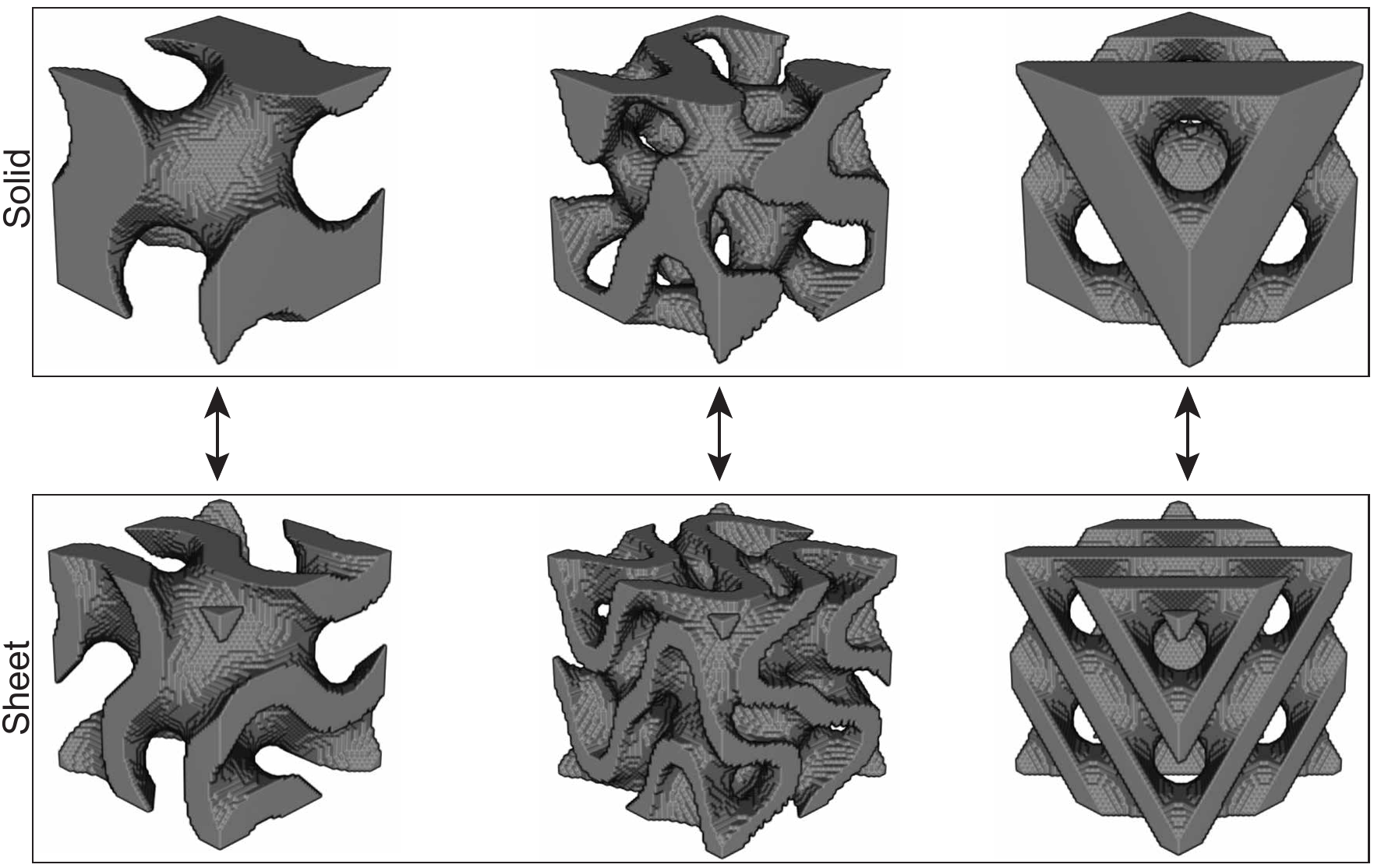}
		\par\end{centering}
	\caption{Six different types of TPMS structures: from left to right, Schoen Gyroid, Fischer Koch S, and Schwarz Diamond. From top to bottom, ``Solid-networks'' and ``Sheet-networks''.\label{fig:6tpms}}
\end{figure}

Our data are 3D TPMS structures at a resolution of 64 in each dimension, with six types as shown in \Cref{fig:6tpms}. Broadly, we classify them as ``Solid-networks'' and ``Sheet-networks''. For each TPMS type, we have 600 3D structures with 64 resolution, totaling 600 data samples per type. The data size is approximately 103 GB, and the data link is \url{https://1drv.ms/f/c/f999b24db97d074b/UgBLB325TbKZIID5YoUEAAAAAK57y3r7MVu56Ds}. The geometries and material distributions are all random. The geometric volume fractions range from 0.26 to 0.66, and Poisson’s ratios range from 0.1 to 0.4. Details of the geometric volume fractions can be found in \ref{sec:TPMS}. Both volume fractions and Poisson’s ratios are uniformly distributed. Given our focus on linear elasticity homogenization, we set Young’s modulus to 1 during training, simplifying the final calculation of the effective elastic tensor by linearly multiplying by Young’s modulus. It is worth noting that our 64-resolution data include the full cube, containing both material and void information; however, we train Transolver using only the point cloud of the material, so the resolution (number of points) varies for each sample.

To test the effect of replay-based continual learning, we first show that the current PDE-trained Transolver performs mediocrely on OOD problems. We divide the experiments into three groups: the first uses ``Solid-networks'' (1800 samples) as the training set; the second uses ``Sheet-networks'' (1800 samples); the third uses both ``Solid-networks'' and ``Sheet-networks'' (3600 samples in total). The results of these three experiments are shown in \Cref{tab:transolver_3D_tpms} in rows 1, 2, and 3. We can draw the same conclusion as for the 2D brain tumors in \Cref{tab:transolver_tumor}: physics-based neural operators do not perform well on OOD problems. Therefore, we need continual learning to address the OOD issue. Next, we test the effect of replay-based continual learning.

We test the replay-based continual learning through four groups of experiments, as shown in rows 4, 5, 6, and 7 of \Cref{tab:transolver_3D_tpms}. Below we describe these four experiments in detail. The first group uses ``Solid-networks'' (1800 samples) as past data and ``Sheet-networks'' (1800 samples) as new data, trained with joint learning. The second group uses ``Sheet-networks'' (1800 samples) as past data and ``Solid-networks'' (1800 samples) as new data, trained with joint learning. The third group uses ``Solid-networks'' (1800 samples) as past data and ``Sheet-networks'' (1800 samples) as new data, trained with replay-based continual learning. The fourth group uses ``Sheet-networks'' (1800 samples) as past data and ``Solid-networks'' (1800 samples) as new data, trained with replay-based continual learning. From the relative errors in rows 4, 5, 6, and 7 of \Cref{tab:transolver_3D_tpms}, we can see that both joint learning and replay-based continual learning achieve good accuracy. Since the amount of new task data in this experiment is large, equal to the amount of past data, the reduction in per-step computation time of replay-based continual learning compared to joint learning is not as significant.
Specifically, since the numbers of new-task and past-task samples are identical in this case, we use all new-task samples and  select 50\% of the past-task samples for replay during continual learning.

\begin{figure}
	\begin{centering}
		\includegraphics[scale=0.55]{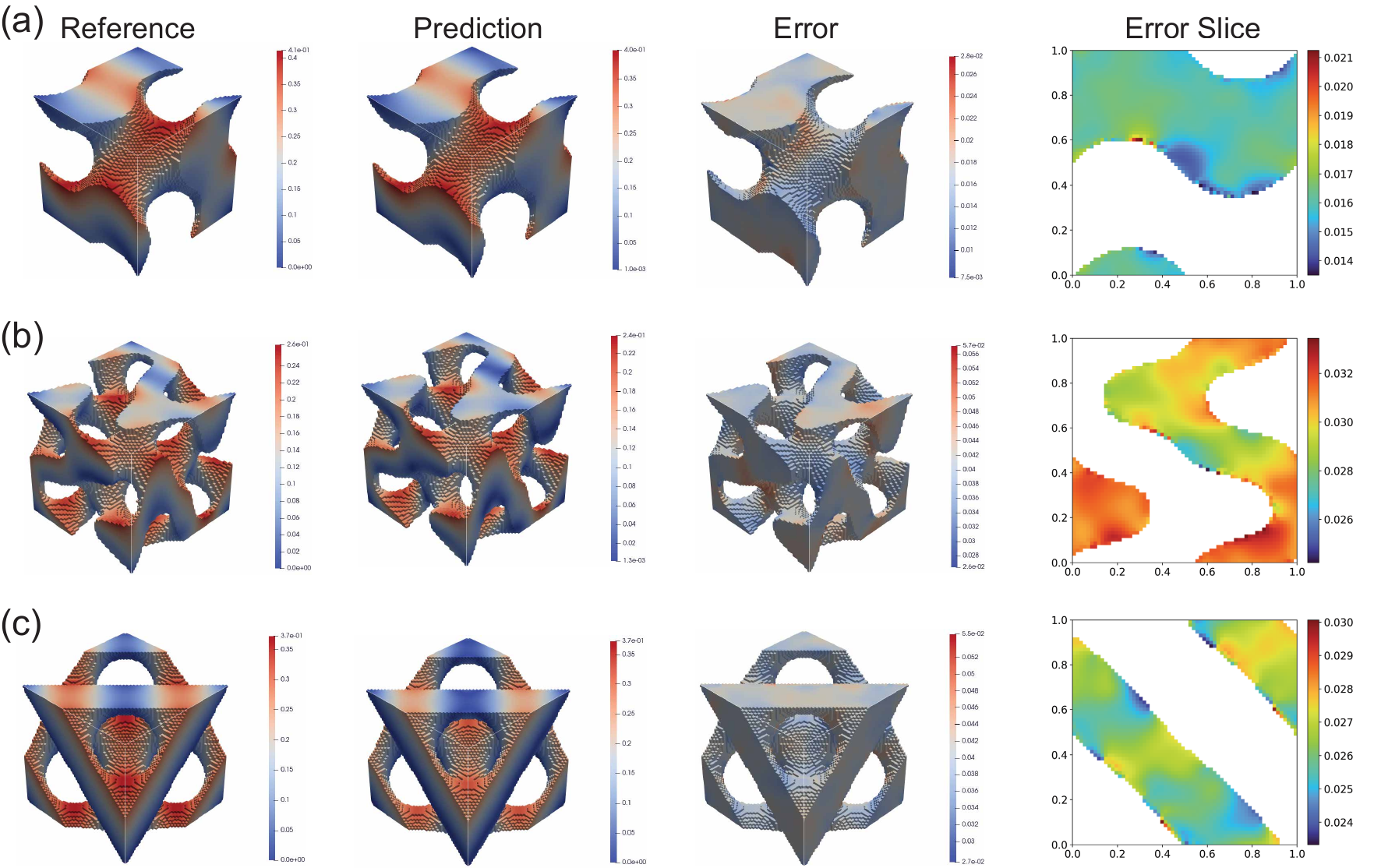}
		\par\end{centering}
	\caption{Prediction of replay-based continual learning with PDE-trained Transolver on ``Solid-networks'' TPMS. From top to bottom: Schoen Gyroid, Fischer Koch S, and Schwarz Diamond. From left to right: FEM reference solution of absolute displacement field, Transolver prediction of absolute displacement field, corresponding absolute displacement error contour, and cross-sectional error contour.\label{fig:6tpms-solid}}
\end{figure}

\begin{figure}
	\begin{centering}
		\includegraphics[scale=0.55]{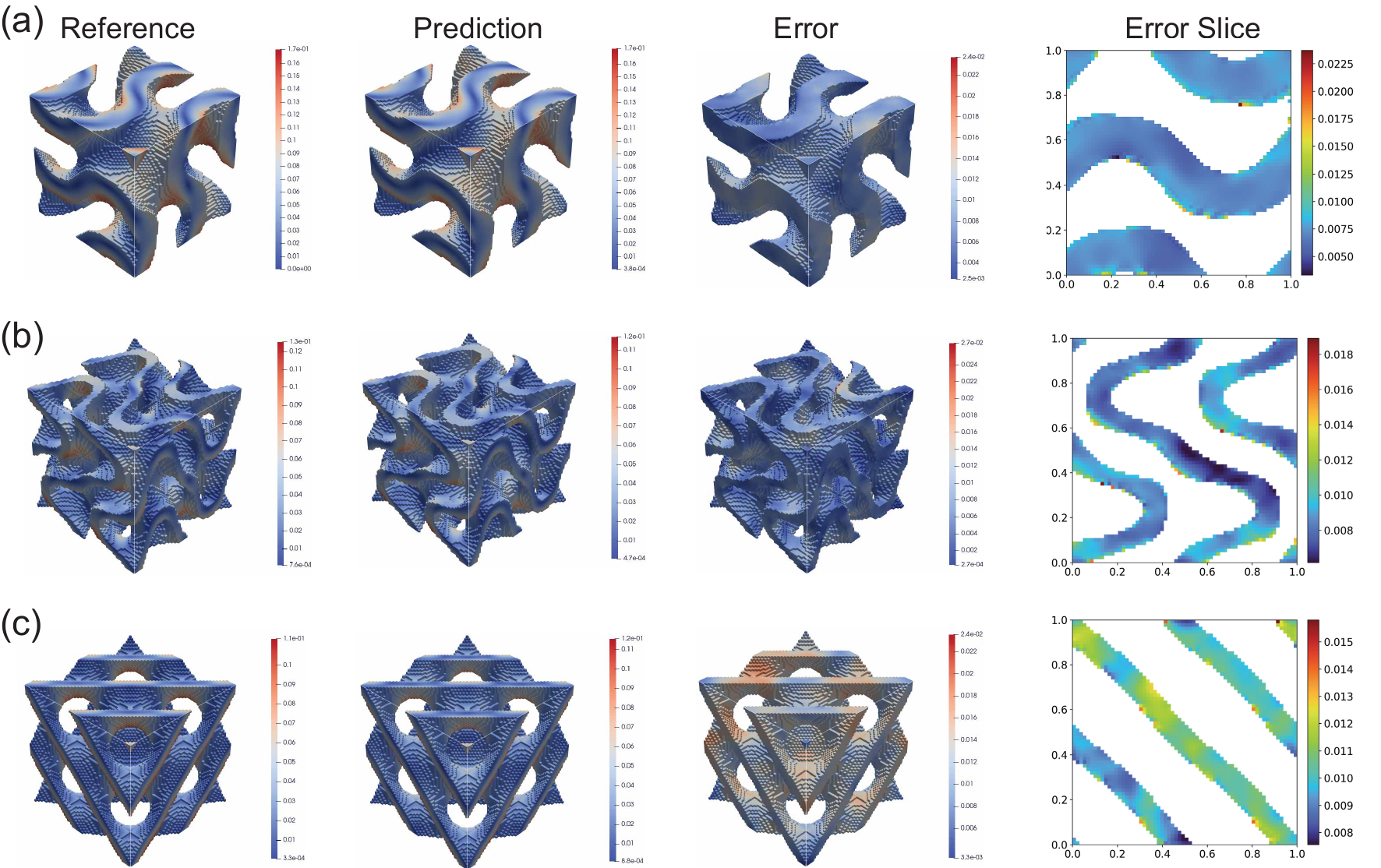}
		\par\end{centering}
	\caption{Prediction of replay-based continual learning with PDE-trained Transolver on ``Sheet-networks'' TPMS. From top to bottom: Schoen Gyroid, Fischer Koch S, and Schwarz Diamond. From left to right: FEM reference solution of absolute displacement field, Transolver prediction of absolute displacement field, corresponding absolute displacement error contour, and cross-sectional error contour.\label{fig:6tpms-sheet}}
\end{figure}

\Cref{fig:6tpms-solid} and \Cref{fig:6tpms-sheet} show the absolute displacement field contours of the x-direction tension predicted by replay-based continual learning on TPMS structures. We can see that replay-based continual learning achieves accuracy comparable to the finite element reference solution on both ``Solid-networks'' and ``Sheet-networks'' structures, with relative errors below $10\%$, while improving efficiency by thousands of times, as shown in \Cref{tab:transolver_3D_tpms}.

\begin{figure}
	\begin{centering}
		\includegraphics[scale=0.55]{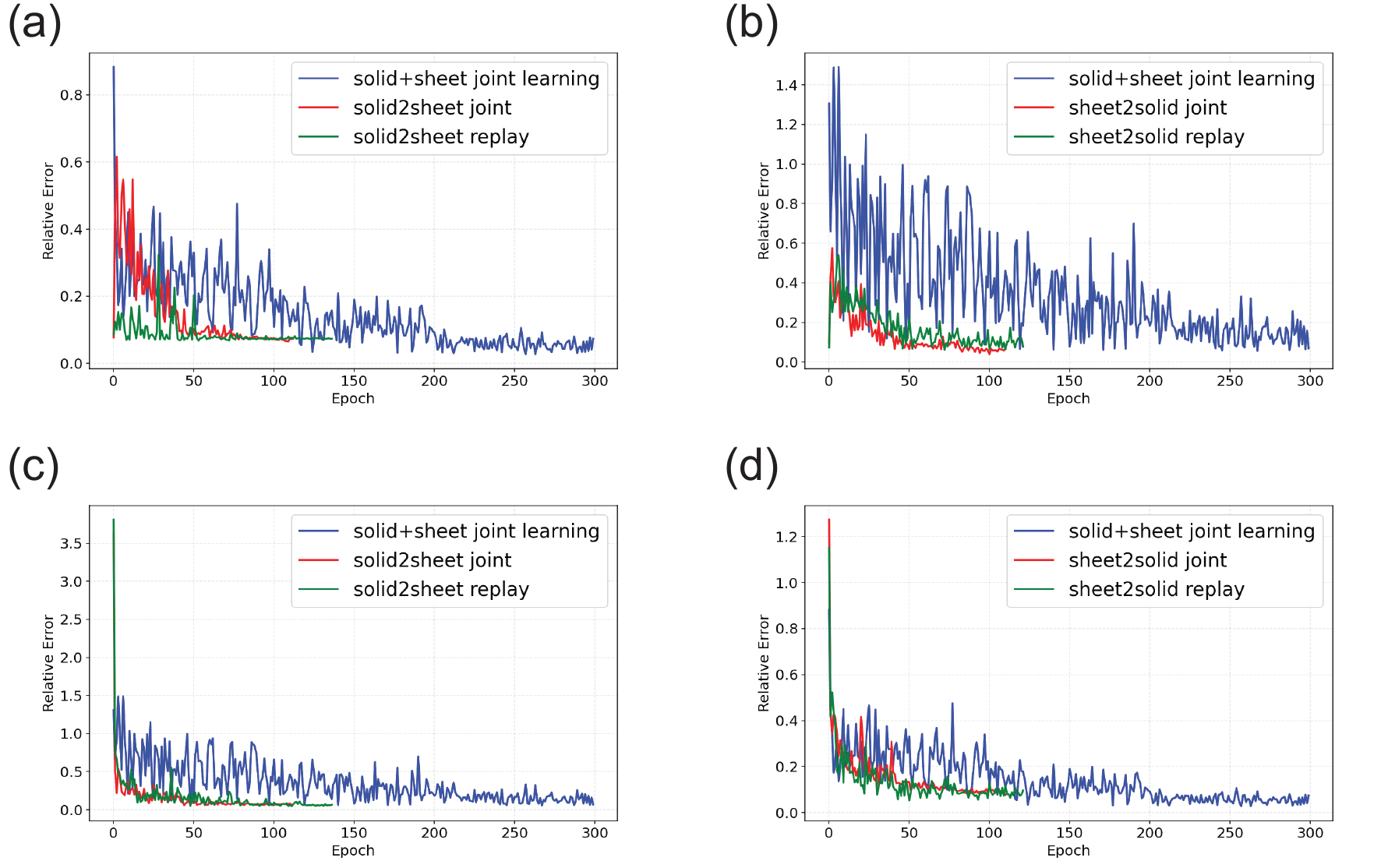}
		\par\end{centering}
	\caption{Variation of relative error with iterations for joint learning and replay-based continual learning on 3D TPMS: (a) past data are ``Solid-networks'', new data are ``Sheet-networks'', relative error on ``Solid-networks''; (b) past data are ``Sheet-networks'', new data are ``Solid-networks'', relative error on ``Sheet-networks''; (c) past data are ``Solid-networks'', new data are ``Sheet-networks'', relative error on ``Sheet-networks''; (d) past data are ``Sheet-networks'', new data are ``Solid-networks'', relative error on ``Solid-networks''. Here, ``solid + sheet joint learning'' means training from scratch by mixing all solid and sheet data; ``solid2sheet joint'' means using the parameters obtained after training on ``Solid-networks'' as the initial model parameters, then performing joint learning on ``Solid-networks'' and ``Sheet-networks''; ``solid2sheet replay'' means using the parameters obtained after training on ``Solid-networks'' as the initial model parameters, then performing replay-based continual learning on ``Solid-networks'' and ``Sheet-networks''.\label{fig:TPMS_relative_error}}
\end{figure}

Next, we further demonstrate the catastrophic forgetting on past data and the adaptation capability on new data of replay-based continual learning during training. \Cref{fig:TPMS_relative_error}a and b show the relative errors of replay-based continual learning on past data. We can see that the relative errors on past data are close to those of joint learning, thus successfully avoiding catastrophic forgetting. Additionally, \Cref{fig:TPMS_relative_error}c and d show the relative errors on new data; we can see that replay-based continual learning achieves relative errors close to those of joint learning, indicating good generalization on new data. Note that the time per epoch of replay-based continual learning is less than that of joint learning, as shown in ``Time: s/epoch'' in \Cref{tab:transolver_3D_tpms}. However, because the amount of new data in this problem is very large, the training efficiency gain of replay-based continual learning is not as high as that for the 2D brain tumor in \Cref{subsec:brain_tumors}.

It is worth noting that the relative error on the past data in \Cref{fig:TPMS_relative_error} initially increases abruptly. This is because, at the early stage of adaptation to the new data, the neural network has not yet found an appropriate balance between the past and new tasks. However, this balance is quickly recovered during training, enabling the model to prevent catastrophic forgetting on the past data while maintaining strong adaptation capability on the new data.

\section{Discussion\label{sec:Discussion}}

\subsection{Benefits of PDE-driven methods compared to data-driven methods\label{subsec:PDEs_benefit}}

\begin{figure}
	\begin{centering}
		\includegraphics[scale=0.8]{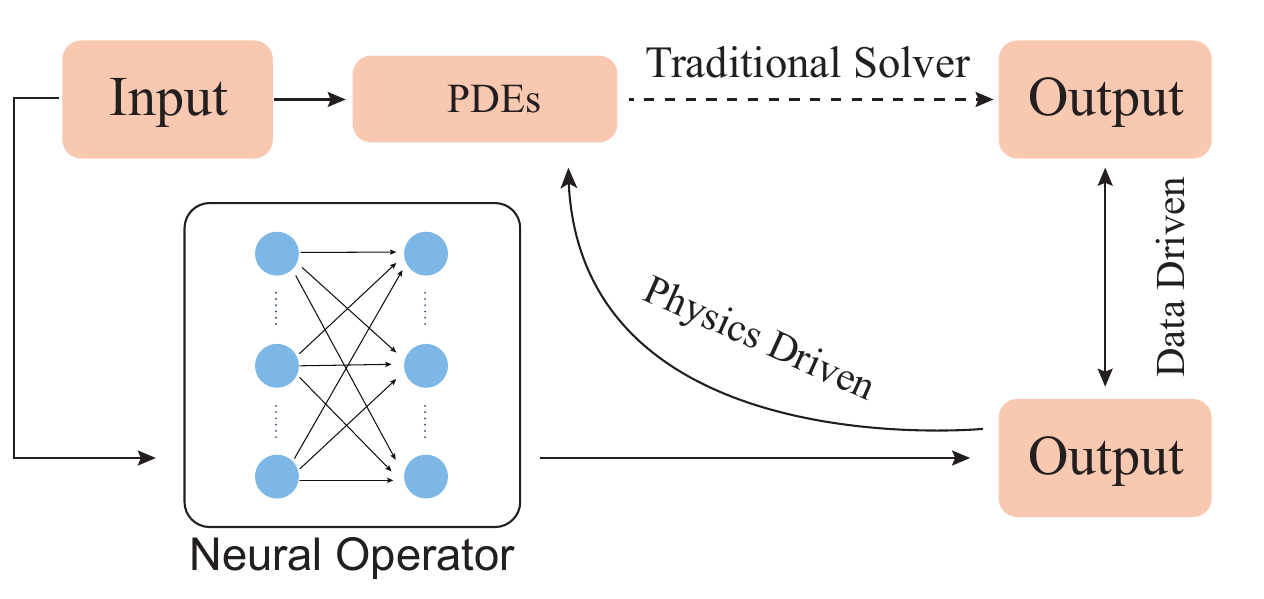}
		\par\end{centering}
	\caption{Comparison of the data-driven and PDE-driven workflows: solid lines indicate the PDE-driven workflow; solid plus dashed lines indicate the data-driven workflow; the dashed line represents the extra steps in the data-driven workflow compared to the PDE-driven one.\label{fig:data_phy_comparision}}
\end{figure}

From the above experiments, we find that on problems where the PDEs are well understood, PDE-driven methods yield significantly greater benefits than data-driven methods when training neural operators. These benefits arise because PDE-driven methods do not require precomputing labels for the corresponding problems, thus saving time on data generation, as shown in \Cref{fig:data_phy_comparision}. In addition, PDE-driven methods can employ explicit differentiation similar to shape functions, which greatly reduces the computational cost of differentiation; typically, the efficiency per epoch can be comparable to that of data-driven methods.

On problems where the PDEs are clear, we strongly recommend PDE-driven methods. In data-driven approaches, conventional solvers are usually used to generate data based on the PDEs, and then neural operators are trained with that data. For some large and complex problems, this is often very time-consuming and unnecessary. Since the source of data-driven methods is the PDEs, the PDEs constitute a complete source of all data. Therefore, it is entirely feasible to train operators using the PDEs themselves. Although PDE-driven methods do not explicitly require data labels, in the process of iterating with PDE information, the neural operator essentially computes the label for each data point. Hence, whether data-driven or PDE-driven, the accuracy on out-of-distribution (OOD) problems will degrade significantly.

\subsection{Closed-loop computational framework}

\begin{figure}
	\begin{centering}
		\includegraphics[scale=0.8]{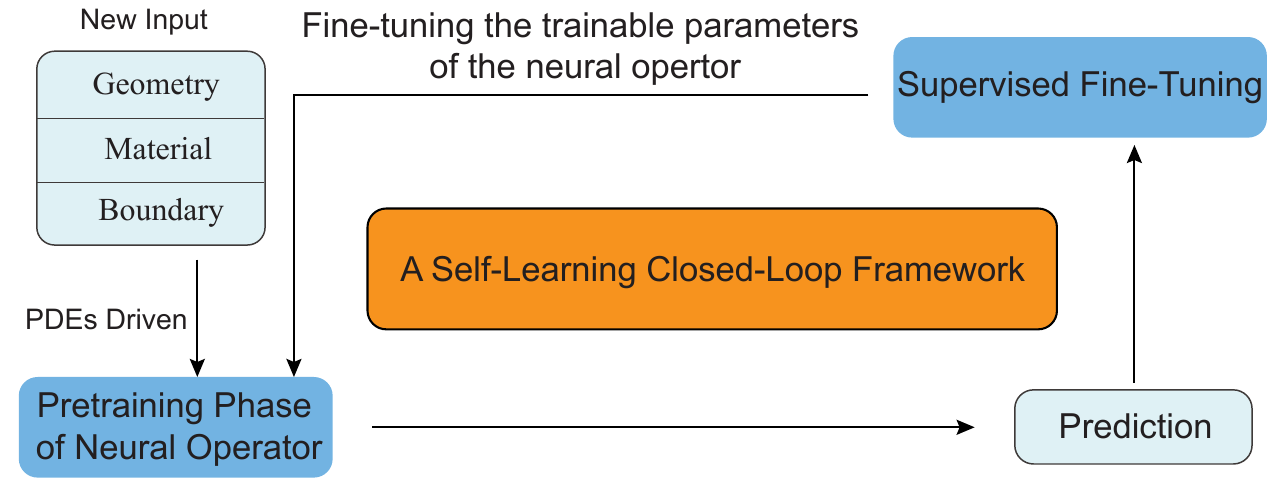}
		\par\end{centering}
	\caption{Schematic diagram of the closed-loop computational framework: geometries, materials, and boundary conditions are processed into point cloud data and fed into the Transolver neural operator, which is then pre-trained using the PDEs. Subsequently, the neural operator undergoes supervised fine-tuning training based on the available labeled data.\label{fig:self_closed_loop}}
\end{figure}

A notable advantage of continual learning is that it can continuously absorb new data at low cost, making the model increasingly powerful. It can also be combined with supervised fine-tuning (SFT) techniques to further improve accuracy. Therefore, broadly speaking, this constitutes a closed-loop computational framework capable of self-learning, as illustrated in \Cref{fig:self_closed_loop}.

Overall, this closed-loop computational framework consists of two stages. The first stage is large-scale pre-training using the PDEs. All data are processed into point clouds, where each point is a high-dimensional information field containing all key simulation information, and then fed into the neural operator. In the manuscript, we adopted Transolver. In the future, new and more powerful neural operator models will emerge; we only need to replace the model with a stronger neural operator, while the overall closed-loop framework shown in \Cref{fig:self_closed_loop} remains unchanged. After large-scale PDE pre-training, if some labeled data are already available, we can use SFT techniques for fine-tuning, similar to the alignment operation in large models, to better adapt to changing scenarios.

\section{Conclusion\label{sec:Conclusion}}

We propose a replay-based continual learning method for neural operators, focusing on physics-driven Transolver. The proposed replay-based continual learning not only significantly reduces catastrophic forgetting on past data, but also exhibits good adaptability on new data. The core idea of the proposed replay-based continual learning is to focus on a small set of poor data, thereby accelerating model updates when incorporating new data. We have extensively validated our method on problems in fluid mechanics (Darcy flow), biology (brain tumor), and solid mechanics (3D TPMS). The results show that the proposed replay-based continual learning can successfully update the model. When the new data are small, the update efficiency of the proposed replay-based continual learning is greatly improved compared to joint learning. The proposed replay-based continual learning also generalizes successfully when the new data are large. In addition, we incorporate supervised fine-tuning (SFT) techniques into neural operators to further enhance accuracy.

Although the proposed replay-based continual learning shows great potential, several key issues and directions merit future research. Currently, the PDE error in the proposed replay-based continual learning is not sufficiently refined; it only directly measures the degree to which the PDEs are satisfied. In the future, we can focus on goal-oriented PDE error estimation for fields of interest, referring to \citet{oden2001goal}. For example, in crack problems, we are usually concerned with the stress intensity factor at the crack tip. Even if the field far from the crack tip deviates significantly from the reference solution, it generally does not affect the stress intensity factor of interest. The region that influences the stress intensity factor is mainly the displacement field near the crack tip; hence, we need to increase the weight around the crack tip to construct a goal-oriented error estimation \citet{oden2001goal}. Moreover, the brain tumor problem is currently based on simulated data; in the future, we will incorporate real MRI data for mechanical analysis, with data available from \url{https://www.cancerimagingarchive.net/collection/ucsf-pdgm/} \citep{calabrese2022university}. At present, we mainly focus on the PDE-driven Transolver model. The proposed replay-based continual learning is theoretically not limited to data-driven or physics-driven approaches, nor to the choice of neural operator model, such as DeepONet or FNO. Future work is warranted to apply and compare the proposed replay-based continual learning on both data-driven and physics-driven methods, as well as on mainstream neural operators.

In the future, large physics models will undoubtedly emerge \citep{menon2026scientific,choi2025defining}. As data increase, it will be necessary to continuously update the parameters of these large physics models. The proposed replay-based continual learning provides an excellent continual learning strategy for handling new data in large physics models, with the key advantages of being very simple while achieving strong performance.

\section*{Declaration of competing interest}
The authors declare that they have no known competing financial interests or personal relationships that could
have appeared to influence the work reported in this paper.

\section*{Acknowledgement}
The study was supported by the Key Project of the National Natural Science Foundation of China (12332005) and scholarship from Bauhaus University in Weimar.  We would like to thank Zhongkai Hao for the insightful discussions.

\section*{CRediT authorship contribution statement}

\textbf{Yizheng Wang}: Conceptualization, Methodology, Formal analysis, Investigation, Data curation, Validation, Visualization, Writing – original draft, Writing – review \& editing. 
\textbf{Mohammad Sadegh Eshaghi}: Conceptualization, Methodology, Investigation.  
\textbf{Xiaoying Zhuang}: Supervision, Writing – review \& editing.  
\textbf{Timon Rabczuk}: Supervision, Writing – review \& editing, Funding acquisition.  
\textbf{Yinghua Liu}: Supervision, Funding acquisition.

\appendix

\section{Input and output of Fourier Neural Operator\label{sec:Input-and-output_FNO}}

\begin{figure}
	\begin{centering}
		\includegraphics[scale=0.65]{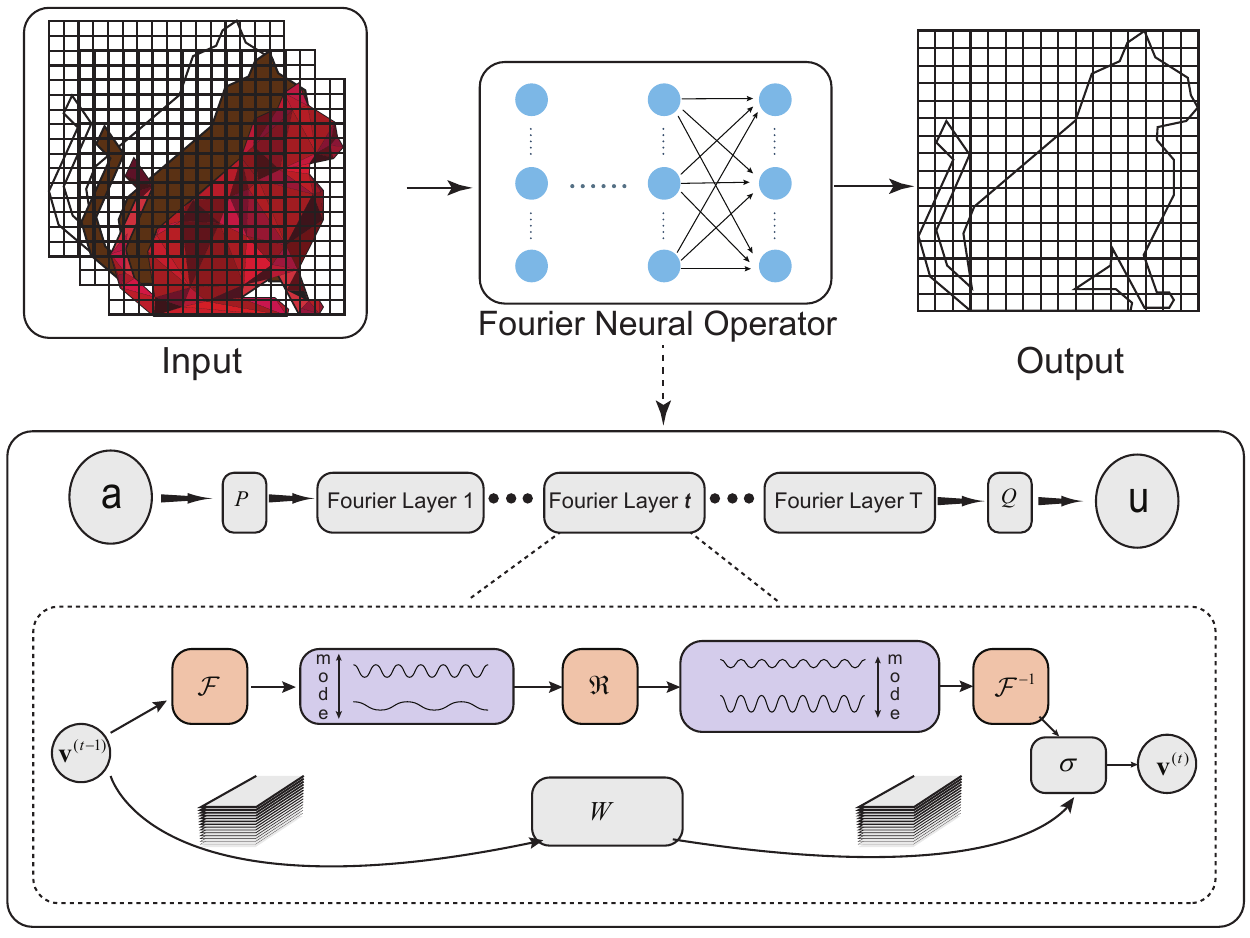}
		\par\end{centering}
	\caption{Input and output of Fourier Neural Operator\label{fig:Input-and-output_FNO}}
\end{figure}

FNO\citet{li2020fourier} is one of the most representative neural operators. The input and output of FNO typically require regular grids, often described using a large matrix, as shown in \Cref{fig:Input-and-output_FNO}. Consider a simple scenario, such as describing a cat-shaped material in \Cref{fig:Input-and-output_FNO}, where material is present (labeled 1) and absent (labeled 0). Clearly, FNO has limited capability in describing complex geometries, which introduces considerable errors. Although some studies aim to improve FNO’s performance on complex geometries, such as Geo-FNO\citet{li2022fourier}, these approaches still face challenges in generalizing to complex geometries.

Therefore, can we abandon the grid-based description of complex geometries? Unfortunately, this is very difficult within the FNO architecture, because the core of the FNO algorithm is the FFT, which achieves high computational efficiency only on regular grid matrices. This limitation arises from the FFT algorithm. For details of FNO, please refer to \citet{li2020fourier}.

\section{Introduction to Transolver\label{sec:transolver_introduction}}

\begin{figure}
	\begin{centering}
		\includegraphics[scale=0.45]{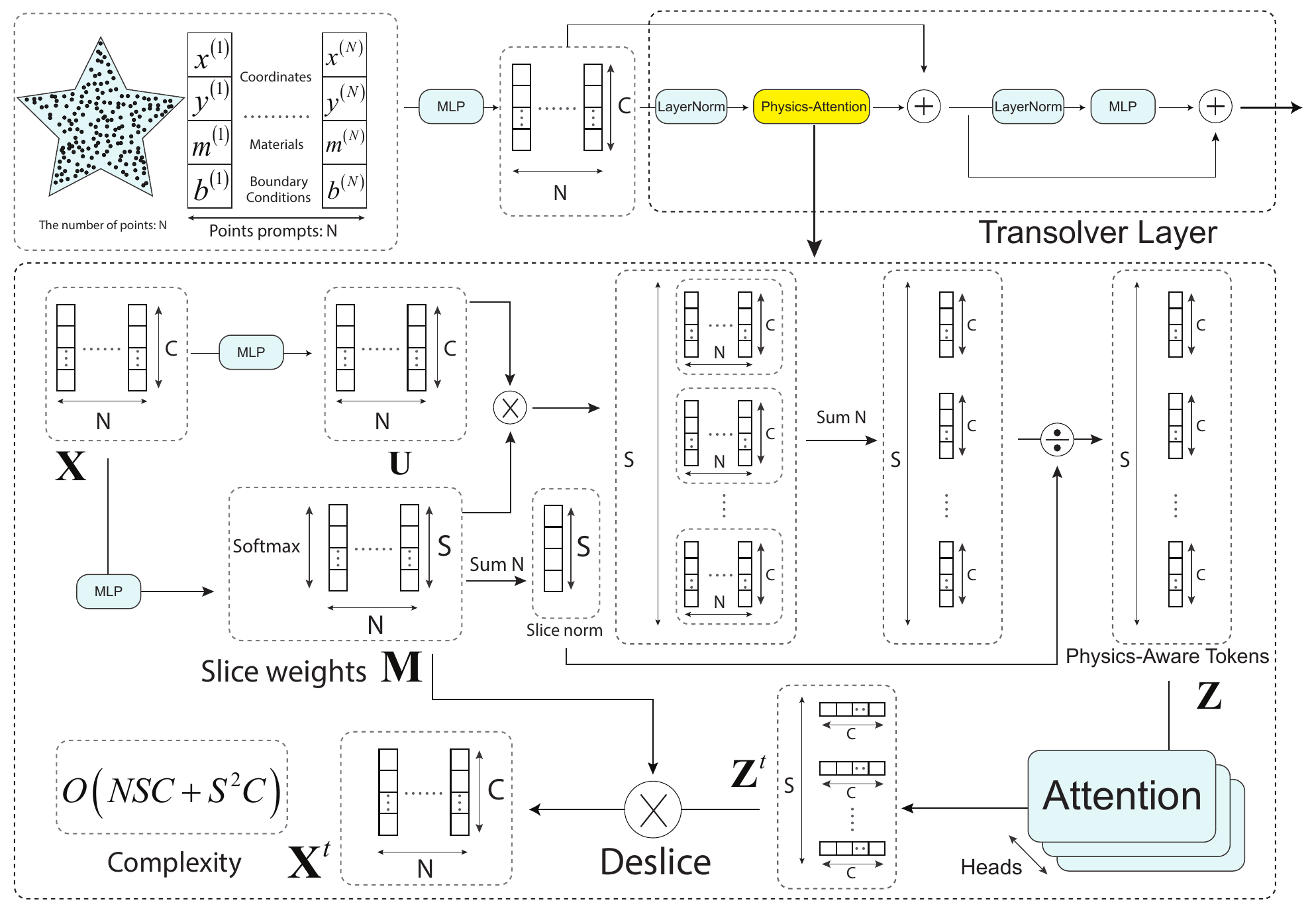}
		\par\end{centering}
	\caption{The architecture of Transolver layer \label{fig:Transolver}}
\end{figure}

The core idea of Transolver is to perform attention on low-complexity physics-aware tokens. The algorithm flow is shown in \Cref{fig:Transolver}. Traditional GNOT\citet{hao2023gnot} performs attention on points, with complexity $O(N^{2})$, where $N$ is the total number of input points. In contrast, Transolver\citet{wu2024transolver} performs attention on physics-aware tokens, reducing the complexity to $O(N+S^{2})$, where $S$ is the number of physics-aware tokens. Since $S \ll N$, Transolver demonstrates great potential for very complex problems.

We explain the algorithm flow of Transolver. First, we encode all points in the domain: $\{x^{(i)}, y^{(i)}, m^{(i)}, b^{(i)}\}^{N}_{i=1}$, where $x^{(i)}$ and $y^{(i)}$ are coordinates, $m^{(i)}$ is the material information at that location (e.g., elastic modulus), and $b^{(i)}$ is the boundary condition at that point. Notably, these points do not need to be regular. These points are passed through an MLP, followed by a Layer Normalization to improve convergence speed by normalizing each feature. The output is $\boldsymbol{X} \in \mathbb{R}^{N \times C}$, where $C$ is the number of channels of point information. Then, the core physics-attention module of Transolver is applied. $\boldsymbol{X}$ is passed through two different MLPs, producing $\boldsymbol{U} \in \mathbb{R}^{N \times C}$ and $\boldsymbol{M} \in \mathbb{R}^{N \times S}$, respectively, where $S$ is the number of physics-aware tokens. The slice weights $\boldsymbol{M}$ are subjected to a softmax operation along the $S$ dimension to give them a weight interpretation. Then, using $\boldsymbol{M}$ as weights, a weighted sum of $\boldsymbol{U}$ is computed to obtain the physics-aware tokens $\boldsymbol{Z} \in \mathbb{R}^{S \times C}$:
\begin{equation}
	Z_{JK} = \frac{\sum^{N}_{I=1} M_{IJ} U_{IK}}{\sum^{N}_{I=1} M_{IJ}}
\end{equation}
Note that the size of $\boldsymbol{Z}$ is much smaller than that of $\boldsymbol{X}$ because $S \ll N$. Then, we apply the multi-head attention mechanism from transformers along the $S$ dimension of $\boldsymbol{Z}$:
\begin{equation}
	\boldsymbol{Q}, \boldsymbol{K}, \boldsymbol{V} = \text{MLP}(\boldsymbol{Z}); \quad \boldsymbol{Z}^{t} = \text{softmax}(\boldsymbol{Q}\boldsymbol{K}^{T})\boldsymbol{V}
\end{equation}
After obtaining $\boldsymbol{Z}^{t} \in \mathbb{R}^{S \times C}$, we use $\boldsymbol{M}$ again to deslice and obtain the final output of physics attention, $\boldsymbol{X}^{t}$:
\begin{equation}
	X^{t}_{IK} = \sum^{S}_{J=1} M_{IJ} Z^{t}_{JK}
\end{equation}

The subsequent operations, as shown in \Cref{fig:Transolver}, are very straightforward and we will not elaborate further. The core of Transolver is the physics-attention module, which greatly reduces the quadratic complexity $O(N^{2})$ of processing points in transformers to a linear complexity $O(N+S^{2})$ for processing physics-aware tokens. At the same time, Transolver has strong approximation capabilities. \citet{wu2024transolver} demonstrated its potential on complex 3D geometry problems, especially on ShapeNet Car. Therefore, Transolver has great research value and promise. It is worth noting that we can stack multiple Transolver layers, similar to Fourier layers in FNO, which improves generalization but also increases computational cost. Depending on the complexity of the problem, we can adjust the number of Transolver layers and the size of $S$. Typically, the more complex the problem, the larger the number of Transolver layers and $S$. $S$ essentially transforms the original point-based description of the problem into a description of physical spatial regions. The core idea of Transolver is that physical problems often exhibit continuous similarities, so the information at each point is redundant.

\section{Introduction to LoRA\label{sec:LoRA_introduction}}

\begin{figure}
	\begin{centering}
		\includegraphics[scale=0.6]{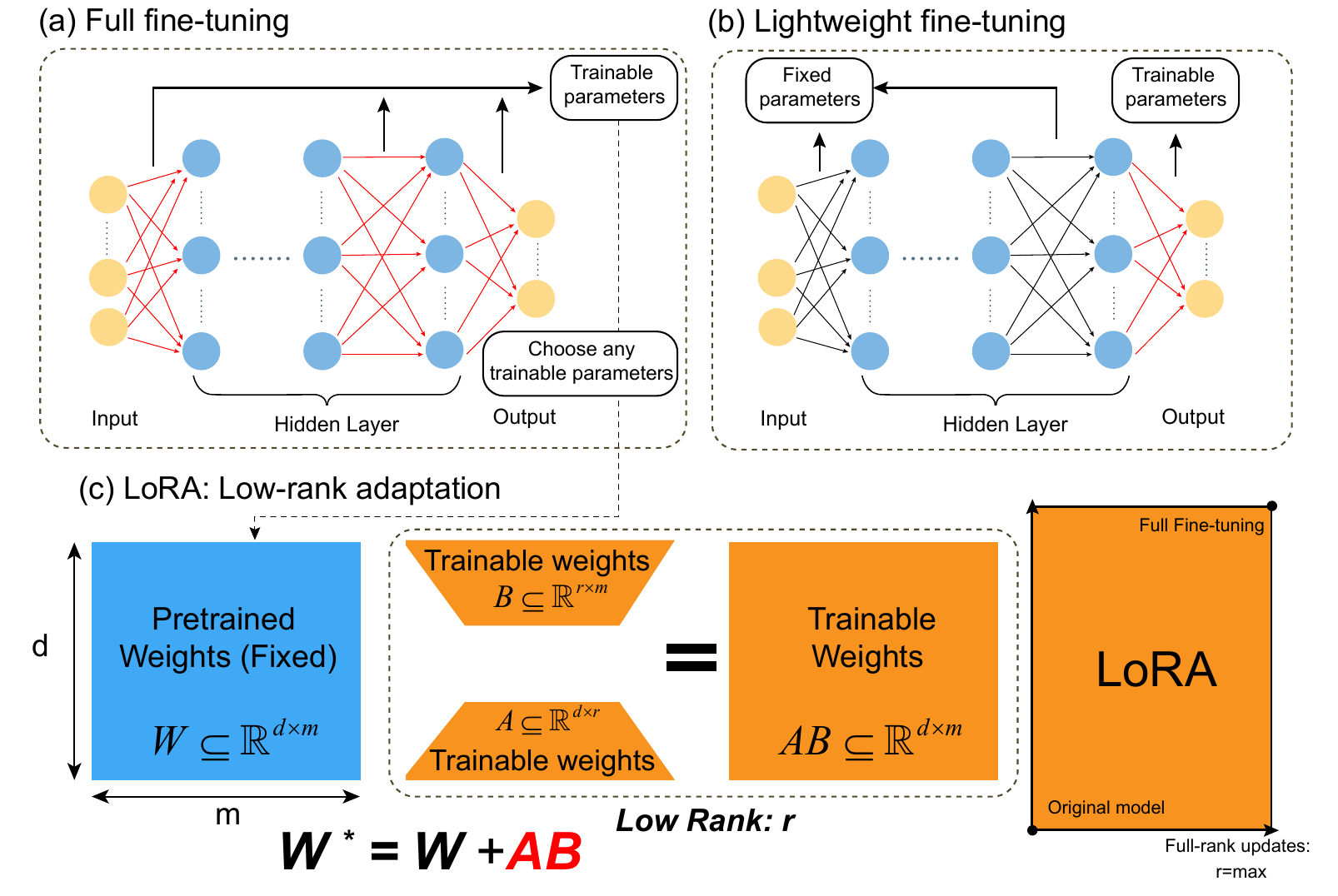}
		\par\end{centering}
	\caption{Three common approaches for parameter-based transfer learning: (a) Full fine-tuning: fine-tune all parameters of the model; the red arrows indicate parameters that need fine-tuning. (b) Lightweight fine-tuning: fine-tune part of the model's parameters; the red arrows indicate parameters that need fine-tuning. (c) LoRA: the blue matrix $\boldsymbol{W}$ is fixed (no training required) and comes from a previously pre-trained model. The yellow matrices $\boldsymbol{A}$ and $\boldsymbol{B}$ are the parameters to be trained in LoRA; $\boldsymbol{AB}$ is a low-rank matrix with rank $r$. $\boldsymbol{A}$ and $\boldsymbol{B}$ are trained on the new dataset. $\boldsymbol{W}^{*}=\boldsymbol{W}+\boldsymbol{A}\boldsymbol{B}$ is the neural network weight during testing.\label{fig:Parameter-based-transfer-learn}}
\end{figure}

Transfer learning has always been a research hotspot in deep learning. Since there are many transfer learning methods, here we focus only on the most common parameter-based transfer learning. Parameter-based transfer learning generally has three approaches: full fine-tuning, lightweight fine-tuning, and LoRA (Low-Rank Adaptation of Large Language Models) \citet{hu2021lora}. The difference between full fine-tuning and lightweight fine-tuning is shown in \Cref{fig:Parameter-based-transfer-learn} a and b.

Given that the most common transfer learning technique currently is LoRA, we focus on introducing LoRA here. The idea of LoRA (Low-Rank Adaptation of Large Language Models) \citet{hu2021lora} is to multiply low-rank matrices $\boldsymbol{AB}$ ($\boldsymbol{A}\subseteq\mathbb{R}^{d\times r}$, $\boldsymbol{B}\subseteq\mathbb{R}^{r\times m}$) and then add them to the original pre-trained parameters $\boldsymbol{W}\subseteq\mathbb{R}^{d\times m}$, as shown in \Cref{fig:Parameter-based-transfer-learn} c. The parameter update formula for LoRA is:

\begin{equation}
	\boldsymbol{W}^{*}=\boldsymbol{W}+\alpha\boldsymbol{A}\boldsymbol{B}
\end{equation}
where $r$ is the rank in LoRA, $r\ll\min(d,m)$. Notably, $\boldsymbol{W}$ is a fixed parameter from the past task and does not require training. We only train $\boldsymbol{A}\boldsymbol{B}$; the number of trainable model parameters in $\boldsymbol{A}\boldsymbol{B}$ is $r*(d+m)$. The number of trainable parameters in $\boldsymbol{A}\boldsymbol{B}$ is far smaller than that in $\boldsymbol{W}$, where the number of trainable parameters in $\boldsymbol{W}$ is $d*m$. $\alpha$ is the scaling factor of LoRA, used to allocate the attention weights between the pre-trained $\boldsymbol{W}$ and the LoRA-trained parameters $\boldsymbol{A}\boldsymbol{B}$; we set it to $16$ by default. $\boldsymbol{A}$ and $\boldsymbol{B}$ are initialized with a Gaussian distribution with mean $0$ and standard deviation $0.02$.

LoRA is essentially a form of lightweight fine-tuning, but more flexible. Moreover, full fine-tuning can be seen as an extreme case of LoRA, i.e., when $r=\min(d,m)$. In summary, the core idea of LoRA is to decompose some weight matrices of the model into the product of two low-rank matrices, thereby reducing the number of parameters that need to be trained.

\section{2D brain tumor data sampling\label{sec:sample_tumors}}

\begin{figure}
	\begin{centering}
		\includegraphics[scale=0.45]{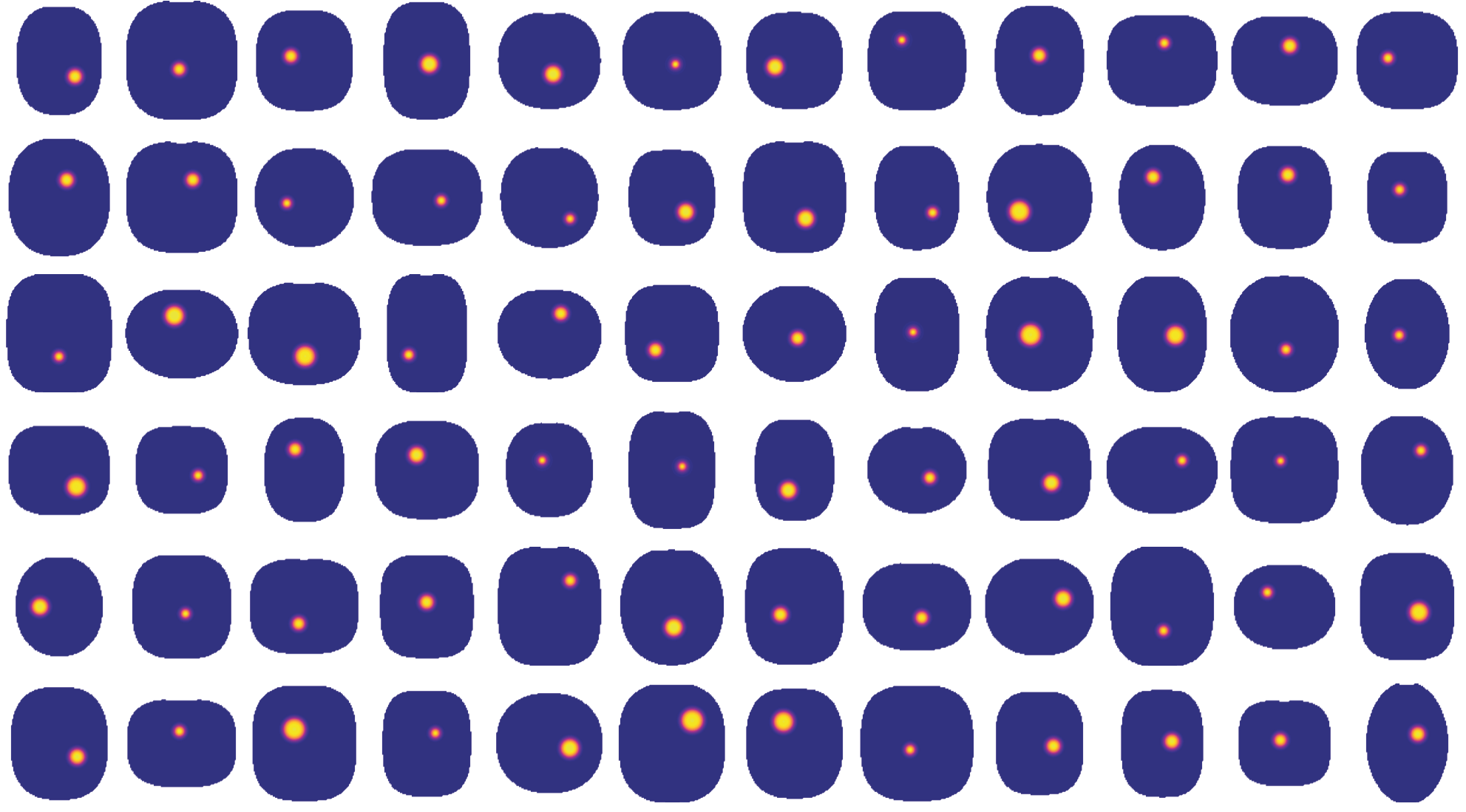}
		\par\end{centering}
	\caption{Sampling of meningioma data\label{fig:menengioma}}
\end{figure}

\begin{figure}
	\begin{centering}
		\includegraphics[scale=0.45]{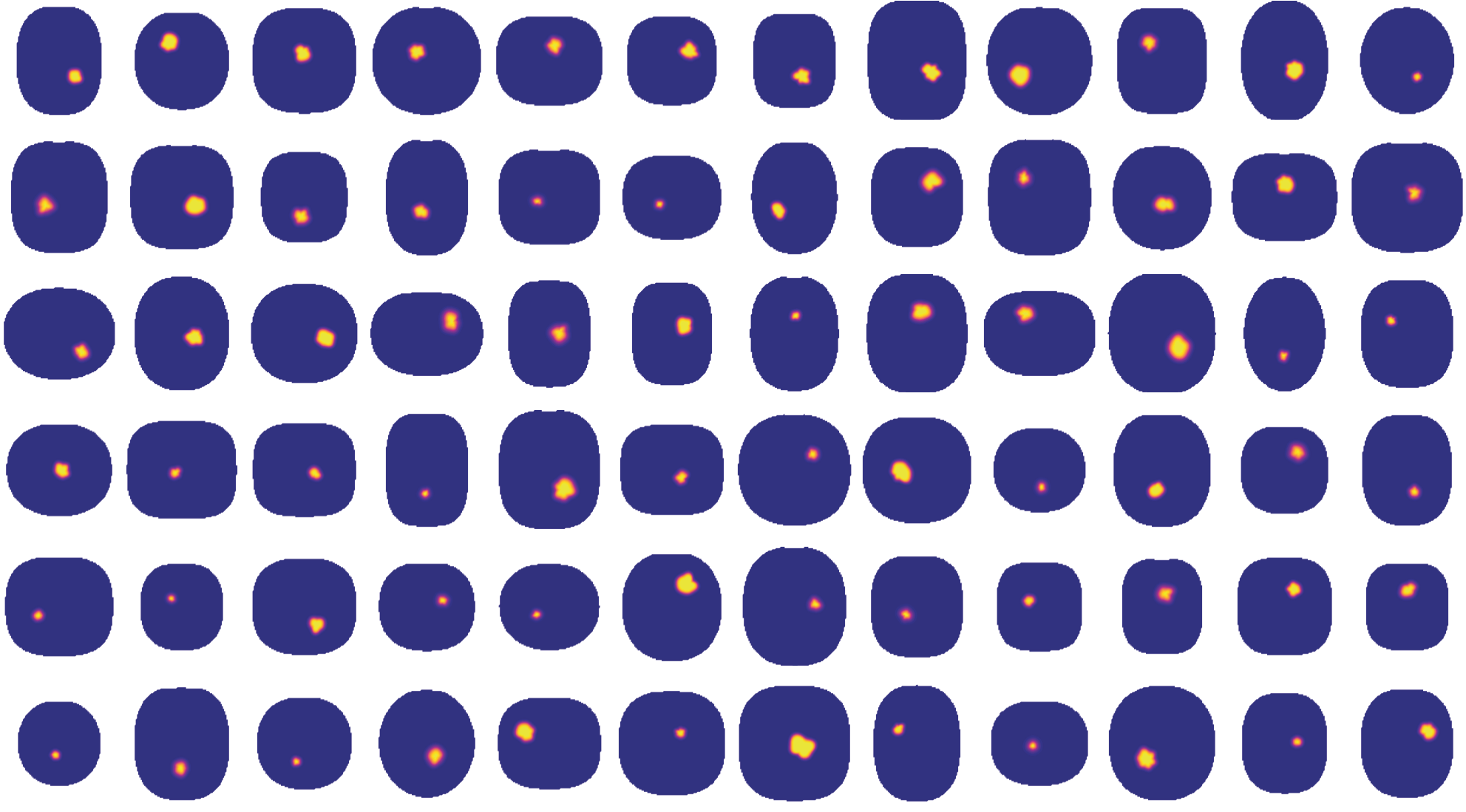}
		\par\end{centering}
	\caption{Sampling of glioma data\label{fig:glioma}}
\end{figure}

We generate 2D brain tumor data based on the shape of the brain and the tumor. Sampling is performed according to random distributions; the specific random sampling parameters are as follows. The left-right width of the brain is $U(0.34,0.48)$, and the top-bottom height is $U(0.36,0.52)$. The radius of the tumor is $U(0.03,0.10)$, and the tumor growth factor is $U(0.06,0.15)$. \Cref{fig:menengioma} and~\Cref{fig:glioma} show sampling for meningioma and glioma. Meningioma is circular, while glioma has an irregular shape.


\section{Numerical homogenization theory\label{sec:homo}}

\begin{figure}
	\begin{centering}
		\includegraphics[scale=0.45]{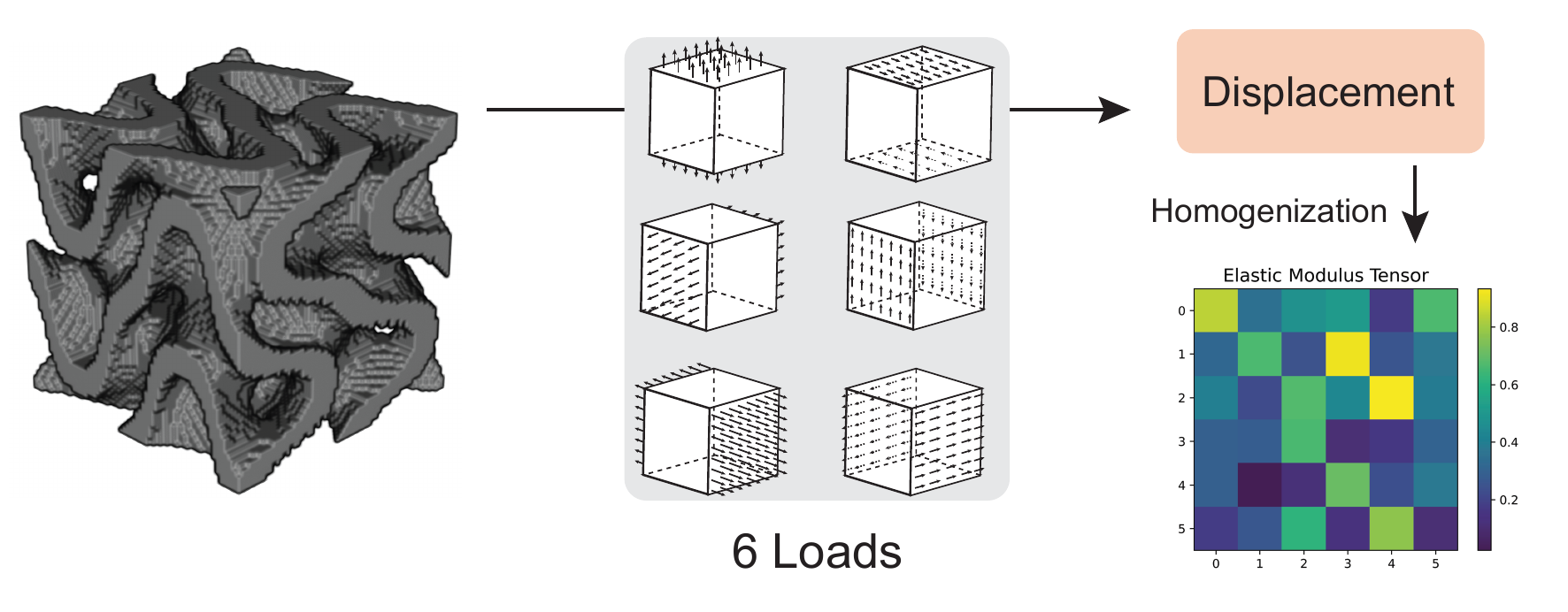}
		\par\end{centering}
	\caption{Schematic diagram of homogenization: a body under six unit strain fields produces six corresponding displacements, and then the displacement responses are processed using the homogenization formula to obtain the elastic tensor.\label{fig:Homogenization_schematic}}
\end{figure}

We introduce the homogenization theory of Andreassen et al. \citet{andreassen2014determine}, which aims to calculate the effective macroscopic elastic tensor of periodic composite materials. We will briefly introduce this method below. For simplicity, we consider a material consisting of solid material and void. $E^{S}_{ijkl}$ denotes the elastic matrix of the solid material. For calculating the homogenized elastic matrix $E^{H}_{IJKL}$, we applied 6 unit strain fields over the unit cell, including three normal strain and three shear strain fields. The 6 unit strain fields $\varepsilon^{(0)(IJ)}_{pq}$ are defined as

\begin{equation}
	\begin{aligned}\boldsymbol{\varepsilon}^{(0)(11)} & =\left[\begin{array}{ccc}
			1 & 0 & 0\\
			0 & 0 & 0\\
			0 & 0 & 0
		\end{array}\right];\boldsymbol{\varepsilon}^{(0)(22)}=\left[\begin{array}{ccc}
			0 & 0 & 0\\
			0 & 1 & 0\\
			0 & 0 & 0
		\end{array}\right];\boldsymbol{\varepsilon}^{(0)(33)}=\left[\begin{array}{ccc}
			0 & 0 & 0\\
			0 & 0 & 0\\
			0 & 0 & 1
		\end{array}\right]\\
		\boldsymbol{\varepsilon}^{(0)(12)} & =\left[\begin{array}{ccc}
			0 & 1 & 0\\
			1 & 0 & 0\\
			0 & 0 & 0
		\end{array}\right];\boldsymbol{\varepsilon}^{(0)(13)}=\left[\begin{array}{ccc}
			0 & 0 & 1\\
			0 & 0 & 0\\
			1 & 0 & 0
		\end{array}\right];\boldsymbol{\varepsilon}^{(0)(23)}=\left[\begin{array}{ccc}
			0 & 0 & 0\\
			0 & 0 & 1\\
			0 & 1 & 0
		\end{array}\right]
	\end{aligned}
	\label{eq:unit strain}
\end{equation}
where $\varepsilon^{(0)(IJ)}_{pq}$ denotes a macroscopic unit strain field imposed on $\Omega$, where the capital indices $IJ$ (as shown in \Cref{eq:unit strain}) correspond to the six different loading cases and do not follow the Einstein summation convention. These six load cases are shown in \Cref{fig:Homogenization_schematic}.

The homogenized elastic matrix $E^{H}_{IJKL}$ is calculated according to the following equation 
\begin{equation}
	E^{H}_{IJKL}=\frac{1}{\vert\Omega\vert}\int_{\Omega}E_{pqrs}(\varepsilon^{(0)(IJ)}_{pq}-\varepsilon^{(v)(IJ)}_{pq})*(\varepsilon^{(0)(KL)}_{rs}-\varepsilon^{(v)(KL)}_{rs})d\Omega\label{eq:homogenized elastic matrix}
\end{equation}
In this equation, $E_{pqrs}$ denotes the local elastic matrix of the TPMS unit cell, and $E_{pqrs}$ obeys 
\begin{equation}
	E_{ijkl}=\begin{cases}
		E^{S}_{ijkl} & \text{for solid material},\\
		0 & \text{for void}.
	\end{cases}
\end{equation}
$\vert\Omega\vert$ represents the volume of the unit cell domain.
$\varepsilon^{(v)(IJ)}_{pq}$ denotes fluctuation strain field calculated by 
\begin{equation}
	\varepsilon^{(v)(IJ)}_{pq}=\varepsilon_{pq}\left(\boldsymbol{X}^{(v)(IJ)}\right)=\frac{1}{2}\left(X^{(v)(IJ)}_{p,q}+X^{(v)(IJ)}_{q,p}\right)
\end{equation}
In this equation, $\boldsymbol{X}^{(v)(IJ)}$ denotes the fluctuation displacement field under applying the macroscopic unit strain field shown in \Cref{eq:unit strain} over $\Omega$. It is worth noting that the total displacement field is expressed as $\boldsymbol{X}^{(IJ)}=\boldsymbol{X}^{(0)(IJ)}-\boldsymbol{X}^{(v)(IJ)}$. Physically, $\boldsymbol{X}^{(0)(IJ)}$ represents the macroscopic affine displacement field corresponding to the uniform strain shown in \Cref{eq:unit strain}, while $\boldsymbol{X}^{(v)(IJ)}$ denotes the fluctuation displacement field induced by the material heterogeneity. We consider the potential energy:
\begin{equation}
	\begin{aligned}\Pi^{(IJ)} & =\int_{\Omega}\frac{1}{2}E_{klpq}\varepsilon^{(IJ)}_{kl}\varepsilon^{(IJ)}_{pq}d\Omega\\
		\varepsilon^{(IJ)}_{kl} & =\frac{1}{2}(X^{(IJ)}_{k,l}+X^{(IJ)}_{l,k})=\frac{1}{2}(X^{(0)(IJ)}_{k,l}-X^{(v)(IJ)}_{k,l}+X^{(0)(IJ)}_{l,k}-X^{(v)(IJ)}_{l,k})=\varepsilon^{(0)(IJ)}_{kl}-\varepsilon^{(v)(IJ)}_{kl}
	\end{aligned}
	\label{eq:homo_energy}
\end{equation}
Taking the first variation of $\Pi^{(IJ)}$ and using $E_{klpq}=E_{pqkl}$, we obtain:
\begin{equation}
	\delta\Pi^{(IJ)}=\int_{\Omega}E_{klpq}\varepsilon^{(IJ)}_{kl}\delta\varepsilon^{(IJ)}_{pq}d\Omega=-\int_{\Omega}E_{klpq}(\varepsilon^{(0)(IJ)}_{kl}-\varepsilon^{(v)(IJ)}_{kl})\delta\varepsilon^{(v)(IJ)}_{pq}d\Omega
\end{equation}
Setting $\delta\Pi^{(IJ)}=0$, $\varepsilon_{pq}(\boldsymbol{X}^{(v)(IJ)})$ is calculated by solving the following weak-form of the equilibrium equation:

\begin{equation}
	\int_{\Omega}E_{klpq}\varepsilon_{kl}(\boldsymbol{X}^{(v)(IJ)*})\varepsilon_{pq}(\boldsymbol{X}^{(v)(IJ)})d\Omega=\int_{\Omega}E_{klpq}\varepsilon_{kl}(\boldsymbol{X}^{(v)(IJ)*})\varepsilon^{(0)(IJ)}_{pq}d\Omega\label{eq:weak_form}
\end{equation}
, where $\boldsymbol{X}^{(v)(IJ)*}$ refers to a virtual fluctuation displacement field. The finite element method is usually used to solve the fluctuation displacement field $\boldsymbol{X}^{(v)(IJ)}$ in the above equation. Then, the fluctuation strain $\boldsymbol{\varepsilon}^{(v)(IJ)}$ can be calculated from $\boldsymbol{X}^{(v)(IJ)}$ and finally substituted into \Cref{eq:homogenized elastic matrix} to solve the homogenized elastic matrix $E^{H}_{ijkl}$.

To ensure a unique solution for the fluctuation displacement field $\boldsymbol{X}^{(v)(IJ)}$ that correctly reflects the material's periodicity, \Cref{eq:weak_form} is solved subject to Periodic Boundary Conditions (PBCs). These conditions require that the fluctuation displacement field $\boldsymbol{X}^{(v)(IJ)}$ takes identical values on opposite faces of the unit cell domain $\Omega$. Assuming $\Omega=[0,L]^{3}$, the PBCs are mathematically expressed as: 
\begin{equation}
	\begin{aligned}\boldsymbol{X}^{(v)(IJ)}(0,x_{2},x_{3}) & =\boldsymbol{X}^{(v)(IJ)}(L,x_{2},x_{3}),\\
		\boldsymbol{X}^{(v)(IJ)}(x_{1},0,x_{3}) & =\boldsymbol{X}^{(v)(IJ)}(x_{1},L,x_{3}),\\
		\boldsymbol{X}^{(v)(IJ)}(x_{1},x_{2},0) & =\boldsymbol{X}^{(v)(IJ)}(x_{1},x_{2},L).
	\end{aligned}
	\label{eq:PBCs}
\end{equation}
where the periodic constraint is essential for the homogenization problem and is enforced in the Finite Element Method implementation by constraining the degrees of freedom of corresponding nodes on opposite faces of the domain. In PFEM, we impose the periodic constraint through a "ring coupling" approach: connecting corresponding nodes on opposite faces with virtual elements and adding the "virtual strain energy" corresponding to these virtual elements to the overall loss function. It should be emphasized that this approach based on virtual strain energy is essentially equivalent to the penalty method; the stiffness matrix of the virtual elements acts as a penalty factor to control the strength of the periodic constraint.

The finite element method is usually used to solve the fluctuation displacement field $\boldsymbol{X}^{(v)(IJ)}$ in \Cref{eq:weak_form}. We observe that the process of homogenization necessitates the computation of displacement fields under six different load conditions. Traditional approaches rely on finite element analysis introduced above, consuming substantial amounts of time for computation. Different geometry and materials should be recalculated to get the corresponding displacement field. Thus, enhancing the speed of calculating displacement fields naturally leads to a significant improvement in the efficiency of the total homogenization process. This is our motivation for using neural operators to improve homogenization problems.

\section{The generation of the TPMS\label{sec:TPMS}}

We introduce the mathematical model of the TPMS. A minimal surface is characterized to have a zero value of mean curvature at any given point of the surface. TPMS is infinitely periodic in three-dimensional space. Level-set approximation is a commonly used approach to represent a minimal surface. The mathematical model of TPMS features level-set equations combined with a series of trigonometric functions, enabling TPMS flexibility to manipulate its topology and mechanical properties. Level-set equations are normally described by 
\begin{equation}
	\phi(x,y,z)=c,
\end{equation}
where $x$, $y$, and $z$ are the spatial coordinates. $\phi(x,y,z)$ on the left-hand side of the equation represents a diffusive field in the 3D space. $\phi=c$ denotes that the level-set value c segments the 3d space of a TPMS unit cell into two connected sub-domains, with each sub-domain representing solid material and void, respectively. The value of c can be used to indicate the volumetric portion of the void. The porosity of the TPMS unit cell can be hence manipulated by adjusting the c value. The iso-surface of TPMS means that an iso-value equals c. 

A broadly utilized level-set equation of TPMS is named Schwarz Primitive formatted by 
\begin{equation}
	\cos(2\pi x)+\cos(2\pi y)+\cos(2\pi z)=c
\end{equation}
The above level-set equation denotes a zero-thickness shell-like topology, known as the minimal surface. We need to further generate load-bearing TPMS lattice material based on the minimal surfaces. Two types of TPMS lattices are formulated based on minimal surfaces. The first type is named "Solid-networks", satisfying 
\begin{equation}
	\phi(x,y,z)>c
\end{equation}
For "solid-networks" TPMS cells, the volume with the diffusive field value $\phi(x,y,z)>c$ is recognized as a solid material, while the rest of the unit cell is recognized as void or pore. The second type is named "sheet-networks". The "sheet-networks" TPMS is governed by 
\begin{equation}
	-c\leq\phi(x,y,z)\leq c
\end{equation}
As its name suggests, "sheet-networks" selects a thin layer of volume around the zero-thickness shell of $\phi(x,y,z)=c$, representing a sheet-like topology of the unit cell. The "solid-networks" and "sheet-networks" TPMS present distinguishable topology characteristics and elastic mechanical properties, offering more flexibility to its "topology-driven" capability.

Another key topological parameter for TPMS is volume fraction (or relative density). Volume fraction refers to the ratio between the volume of solid material and the volume of the entire TPMS unit cell. The volume fraction is directly related to the effective elastic tensor and stiffness of a TPMS unit cell and hereby affects the homogenization elastic matrix. 

In this study, we created a series of TPMS unit cells for training and verifying the proposed neural operator-based homogenization method. In this study, we selected 3 different TPMS level-set equations, including "Schoen Gyroid", "Schwarz Diamond", and "Fischer Kosh S". The level-set equations for each unit cell type are listed: 
\begin{equation}
	\begin{aligned} & SchoenGyroid:\sin(2\pi x)\cos(2\pi y)+\sin(2\pi y)\cos(2\pi z)\\
		& \qquad\qquad\qquad+\sin(2\pi z)\cos(2\pi x)=c\\
		& SchwarzDiamond:\sin(2\pi x)\sin(2\pi y)\sin(2\pi z)\\
		& \qquad\qquad\qquad+\sin(2\pi x)\cos(2\pi y)\cos(2\pi z)\\
		& \qquad\qquad\qquad+\cos(2\pi x)\sin(2\pi y)\cos(2\pi z)\\
		& \qquad\qquad\qquad+\cos(2\pi x)\cos(2\pi y)\sin(2\pi z)=c\\
		& FischerKoshS:\cos(4\pi x)\sin(2\pi y)\cos(2\pi z)+\\
		& \qquad\qquad\qquad\cos(2\pi x)\cos(4\pi y)\sin(2\pi z)\\
		& \qquad\qquad\qquad+\sin(2\pi x)\cos(2\pi y)\cos(4\pi z)=c
	\end{aligned}
	\label{eq:TPMStype}
\end{equation}

\section{Supplementary code}
The code of this work will be available at \url{https://github.com/yizheng-wang/Research-on-Solving-Partial-Differential-Equations-of-Solid-Mechanics-Based-on-PINN} after accepted.

\bibliographystyle{elsarticle-num}
\addcontentsline{toc}{section}{\refname}\bibliography{reference.bib}

\end{document}